\documentclass[]{style}

\usepackage[toc,page,header]{appendix}

\usepackage{minitoc}
\usepackage{subcaption}
\usepackage{pifont}
\usepackage{xcolor,colortbl}
\usepackage{amsfonts}
\usepackage{caption}
\usepackage{float}
\usepackage{tabularx}
\usepackage{mathtools}

\usepackage{footmisc}
\usepackage{algpseudocode}
\usepackage{amssymb}
\usepackage{bm}
\usepackage{booktabs,siunitx,multirow, pifont}
\usepackage[dvipsnames]{xcolor}
\usepackage{siunitx}
\usepackage{tabularx}
\usepackage{makecell}
\usepackage{float}
\usepackage[table]{xcolor}
\usepackage{placeins} 
\usepackage[most]{tcolorbox}
\usepackage{wrapfig}
\usepackage[misc]{ifsym}

\usepackage{makecell}

\definecolor{w_blue}{RGB}{52,204,204}
\definecolor{w_yellow}{RGB}{255,192,0}

\usepackage{xspace}

\usepackage{colortbl}
\usepackage{multicol}

\definecolor{vla_purple}{RGB}{138,83,192}
\definecolor{vla_green}{RGB}{78,167,46}

\definecolor{tpami_blue}{RGB}{52,204,204}
\definecolor{tpami_gray}{RGB}{165,165,165}
\definecolor{tpami_red}{RGB}{192,0,0}
\definecolor{tpami_yellow}{RGB}{248,190,50}
\definecolor{link}{RGB}{248,190,50}

\newcommand{\crbx}[2]{\scalebox{0.75}{\fcolorbox{#1}{#1}{\makebox[1.5ex][c]{\rule{0pt}{1.5ex}#2}}}}

\usepackage{titletoc}

\newcommand{\iconImg}{\raisebox{-0.2ex}{\includegraphics[width=0.018\linewidth]{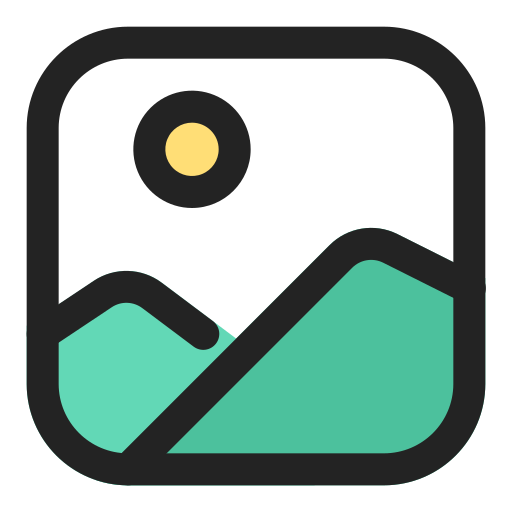}}}
\newcommand{\iconVid}{\raisebox{-0.2ex}{\includegraphics[width=0.018\linewidth]{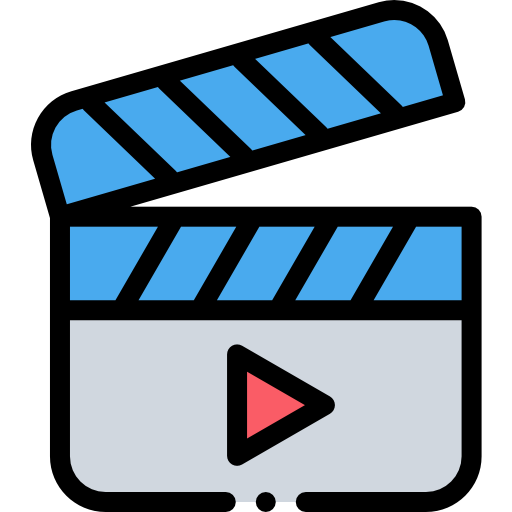}}}

\definecolor{crArgoverse}{RGB}{220,100,180}

\definecolor{crBDD100}{RGB}{150,170,220}
\definecolor{crBDDX}{RGB}{110,180,200}
\definecolor{crBenchDrive}{RGB}{210,170,115}
\definecolor{crDriveBench}{RGB}{130,140,90}

\definecolor{crCarla}{RGB}{230,170,120}
\definecolor{crCarlaSC}{RGB}{150,140,110}
\definecolor{crCoVLA}{RGB}{240,210,100}
\definecolor{crCamOcc}{RGB}{220,140,120}
\definecolor{crOpenDV}{RGB}{235,140,120}

\definecolor{crDriveMLM}{RGB}{210,160,170}
\definecolor{crDriveLM}{RGB}{110,110,130}
\definecolor{crDriveCoT}{RGB}{180,140,120}
\definecolor{crDriveAction}{RGB}{140,150,110}
\definecolor{crDragBench}{RGB}{180,120,200}

\definecolor{crHBD}{RGB}{200,220,80}

\definecolor{crEdiVal}{RGB}{200,150,100}

\definecolor{crGobja}{RGB}{233,30,99}        

\definecolor{crImpromptuVLA}{RGB}{245,170,190}
\definecolor{crInsViE}{RGB}{120,190,170}
\definecolor{crIntern}{RGB}{255,152,0}       

\definecolor{crnuScenes}{RGB}{100,170,150}
\definecolor{crnuPlan}{RGB}{170,200,120}
\definecolor{crNoc}{RGB}{180,170,140}
\definecolor{crNAVSIM}{RGB}{140,160,180}
\definecolor{crNuInstruct}{RGB}{220,120,140}
\definecolor{crNuInteract}{RGB}{190,90,130}

\definecolor{crOpenOcc}{RGB}{170,140,190}
\definecolor{crOccThreeD}{RGB}{220,130,110}
\definecolor{crOmniDrive}{RGB}{130,180,130}
\definecolor{crOmniReasonN}{RGB}{100,150,110}
\definecolor{crOmniReasonB}{RGB}{160,190,220}
\definecolor{crObja}{RGB}{0,188,212}         
\definecolor{crOmni}{RGB}{121,85,72}         
\definecolor{crObjaXL}{RGB}{96,125,139}      

\definecolor{crPrivate}{RGB}{100,100,100}
\definecolor{crProcGen}{RGB}{90,130,130}
\definecolor{crPhysicalAI}{RGB}{100,120,32}
\definecolor{crPanda}{RGB}{15,157,88}        

\definecolor{crRoboBEV}{RGB}{200,110,80}
\definecolor{crReasonDrive}{RGB}{100,150,110}

\definecolor{crSDN}{RGB}{90,140,180}
\definecolor{crSUPAD}{RGB}{190,140,210}

\definecolor{crTEdBench}{RGB}{100,170,200}
\definecolor{crTalkCar}{RGB}{140,90,170}
\definecolor{crTextShape}{RGB}{156,39,176}   

\definecolor{crVLAAD}{RGB}{240,180,80}

\definecolor{crWaymo}{RGB}{135,180,225}
\definecolor{crWOMDR}{RGB}{220,100,100}
\definecolor{crWODEE}{RGB}{100,220,150}

\definecolor{crLMDrive}{RGB}{160,180,100}
\definecolor{crLyft}{RGB}{110,160,120}
\definecolor{crLaionFour}{RGB}{66,133,244}   
\definecolor{crLaionFive}{RGB}{13,71,161}    

\definecolor{crMagicBrush}{RGB}{230,140,90}
\definecolor{crMetaAD}{RGB}{190,120,120}
\definecolor{crCoco}{RGB}{219,68,55}         

\definecolor{crDataset}{HTML}{AFCBFF}  
\definecolor{crBenchmark}{HTML}{E8CFA0} 

\newcommand{\bdgDataset}{\crbx{crDataset}{\textcolor{white}{\textbf{\textsf{D}}}}}
\newcommand{\bdgBenchmark}{\crbx{crBenchmark}{\textcolor{white}{\textbf{\textsf{B}}}}}

\usepackage{tabularx} 
\usepackage{enumitem} 

\usepackage{forest}
\usepackage{algorithm}
\usepackage{listings}
\usepackage{amsmath}

\title{LMMs Meet Object-Centric Vision: Understanding, Segmentation, Editing and Generation}

\author[1\,*]{Yuqian Yuan}
\author[1\,*,\S]{Wenqiao Zhang}
\author[1\,*]{Juekai Lin}
\author[1\,*]{Yu Zhong} 
\author[1\,*]{Mingjian Gao}
\author[1\,*]{Binhe Yu}
\author[1\,*]{Yunqi Cao}
\author[2]{Wentong Li}
\author[1]{Yueting Zhuang}
\author[1]{Beng Chin OOI}

\affiliation[1]{Zhejiang University\\}
\affiliation[2]{Nanjing University of Aeronautics and Astronautics}

\contribution{These authors contributed equally to this research.}

\abstract{Large Multimodal Models (LMMs) have achieved remarkable progress in general-purpose vision--language understanding, yet they remain limited in tasks requiring precise object-level grounding, fine-grained spatial reasoning, and controllable visual manipulation. In particular, existing systems often struggle to identify the correct instance, preserve object identity across interactions, and localize or modify designated regions with high precision. Object-centric vision provides a principled framework for addressing these challenges by promoting explicit representations and operations over visual entities, thereby extending multimodal systems from global scene understanding to object-level understanding, segmentation, editing, and generation. This paper presents a comprehensive review of recent advances at the convergence of LMMs and object-centric vision. We organize the literature into four major themes: object-centric visual understanding, object-centric referring segmentation, object-centric visual editing, and object-centric visual generation. We further summarize the key modeling paradigms, learning strategies, and evaluation protocols that support these capabilities. Finally, we discuss open challenges and future directions, including robust instance permanence, fine-grained spatial control, consistent multi-step interaction, unified cross-task modeling, and reliable benchmarking under distribution shift. We hope this paper provides a structured perspective on the development of scalable, precise, and trustworthy object-centric multimodal systems.}

\checkdata[Correspondence $\S$]{Wenqiao Zhang}

\checkdata[GitHub Repo]{\url{https://github.com/DCDmllm/Awesome-Object-Centric-LMMs}}

\begin{document}
{%
\renewcommand\twocolumn[1][]{#1}%
\maketitle
}

\section{Introduction}
\label{sec:intro}

Large multimodal models (LMMs) have recently emerged as a general-purpose interface between vision and language, demonstrating remarkable capabilities across a broad range of vision-language tasks, including image understanding, visual question answering, instruction following, and multimodal reasoning~\cite{bai2023qwen,bai2025qwen3,guo2025seed1,wang2025internvl3,liu2023visual,zhang2024hyperllava,lin2026mmfinereason,lin2026scientific}. By scaling model capacity, training data, and cross-modal alignment, LMMs have substantially advanced the unified modeling of heterogeneous modalities and reshaped visual intelligence into an increasingly instruction-driven paradigm.

Despite this rapid progress, LLM-based \textbf{object-level intelligence} remains a fundamental bottleneck. In many real-world scenarios, a model must go beyond holistic scene description and operate on \textbf{specific objects} with precision. This requires the ability to (i) identify the correct instance among multiple similar candidates, (ii) reason about object attributes, spatial relations, geometry, and occlusion, and (iii) support fine-grained and controllable actions, such as segmenting a referred object, editing a designated region without affecting others, or generating images that satisfy explicit object-level constraints. 
Existing LMMs, however, typically 
suffer from mis-grounding, weak spatial precision, and limited object permanence in multi-turn interactions. These limitations point to a deeper mismatch between high-level language reasoning and the pixel- or region-level representations required for reliable object-centric perception and manipulation.

\begin{figure*}[t]
\centering
\includegraphics[width=\textwidth]{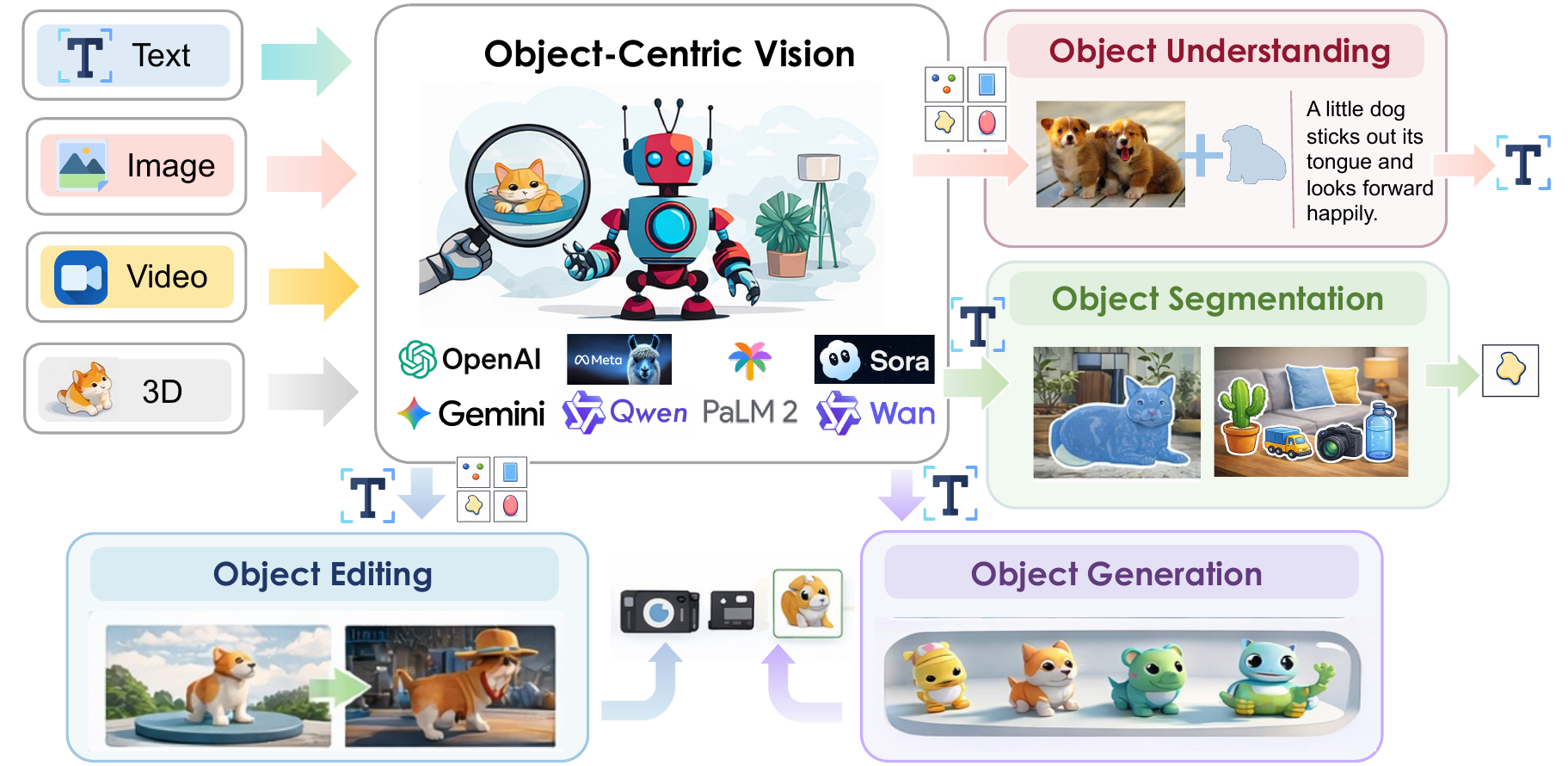}
\caption{
\textbf{Overview of object-centric vision in the era of large multimodal models.}
The modalities considered in this paper include images, videos, and 3D data. Object-centric representations enable four fundamental capabilities: object understanding, object segmentation, object editing, and object generation.
}
\label{fig:intro}
\end{figure*}

Object-centric vision provides a principled perspective for bridging this gap. Rather than treating an image as an undifferentiated array of visual features, object-centric approaches emphasize the discovery, representation, grounding, and manipulation of discrete entities---instances with boundaries, attributes, identities, and relations. When integrated with LMMs, this perspective shifts multimodal systems from passive scene description toward more actionable and controllable capabilities: identifying objects, segmenting them, editing them, and generating them under explicit object-level constraints. Such capabilities are central to a wide range of applications, including interactive agents~\cite{fan2024videoagent,xin2026agentvln}, embodied intelligence~\cite{dang2025rynnec,dang2026rynnbrain,yuan2025eoc,hu2025vision,wang2025forging,xin2026decovln}, 4D reasoning~\cite{huang2026thinking,yin2026mllm},  
visual design~\cite{brooks2023instructpix2pix,wang2024instancediffusion} and medical analysis~\cite{lin2025healthgpt}, where object faithfulness, spatial accuracy, and controllability are essential.

In this paper, we organize the emerging landscape of \textbf{object-centric vision in the era of large multimodal models} into four closely related research themes.

\textbf{Object-Centric Visual Understanding} forms the foundation. It concerns how models perceive, ground, and reason about object instances, attributes, spatial relations, interactions, and scene structure across images, videos, and 3D data. Beyond global scene recognition, this line of research emphasizes fine-grained and grounded understanding that links language to specific visual entities.

\textbf{Object-Centric Referring Segmentation} extends object understanding to precise delineation. Given linguistic or multimodal references, the goal is to identify and segment the target object at the pixel level. With the rise of promptable and open-vocabulary segmentation models, this field has evolved from category-closed supervised learning to more flexible paradigms that support text-guided, interactive, and generalizable object extraction.

\textbf{Object-Centric Visual Editing} studies how to manipulate specific objects or regions while preserving the remaining visual content and overall realism. This includes instruction-guided editing, region-aware modification, subject-preserving transformation, and compositional control. Compared with generic image editing, the object-centric setting places greater emphasis on faithful grounding, spatial accuracy, and controllable local changes.

\textbf{Object-Centric Visual Generation} concerns the synthesis of novel visual content under explicit object-level conditions, such as object identity, layout, attributes, relations, or reference exemplars. Recent advances in diffusion and autoregressive multimodal generation have enabled increasingly strong controllability, making it possible to synthesize images that are not only photorealistic but also structurally and semantically aligned with object-centric constraints.

Although we discuss these four themes separately for clarity, they are deeply interconnected. Object understanding provides the semantic and structural foundation for segmentation, editing, and generation. Segmentation offers explicit object decomposition that facilitates precise manipulation and compositional synthesis. Editing lies at the intersection of understanding and generation, requiring both accurate grounding and high-fidelity rendering. Generation, in turn, supports the other tasks through controllable data synthesis, augmentation, and simulation. Recent research suggests a broader trend of convergence, in which unified architectures, shared representations, and interactive feedback loops integrate understanding, segmentation, editing, and generation into a common object-centric multimodal framework.

Against this background, this paper provides a comprehensive and structured review of object-centric vision in the age of LMMs. Specifically, our contributions are threefold. First, we present a systematic taxonomy spanning object-centric understanding, referring segmentation, visual editing, and visual generation, organizing a rapidly growing body of literature into a coherent conceptual framework. Second, we analyze the technical foundations underlying these capabilities, including multimodal alignment, grounded representation learning, promptable perception, and controllable generation mechanisms. Third, we discuss open challenges and future opportunities, such as improving grounding accuracy, spatial reasoning, fine-grained controllability, cross-task unification, and scaling object-centric modeling from images to videos and 3D environments. 
We further emphasize an object-centric perspective that highlights the shared grounding requirements, common technical challenges, and emerging unification of these tasks under LMMs.

The remainder of this paper is organized as follows. Section~\ref{sec2:background} presents the background and preliminaries of object-centric vision in the era of large multimodal models. Section~\ref{sec3:understanding} begins with object-centric visual understanding, which lays the foundation for grounding and reasoning about visual entities. Building on this basis, Section~\ref{sec4:segmentation} examines object-centric referring segmentation, moving from semantic understanding to precise object-level localization. Section~\ref{sec5:editing} then turns to object-centric visual editing, where grounded understanding enables the controllable manipulation of designated objects or regions. Section~\ref{sec6:generation} further extends this paradigm to object-centric visual generation, with a focus on synthesizing content under explicit object-level constraints. Section~\ref{sec7:future} discusses open challenges and promising directions toward more unified, precise, and reliable object-centric multimodal systems. Finally, Section~\ref{sec8:conclusion} concludes the paper.

\begin{figure*}[]
\centering
\includegraphics[width=\textwidth]{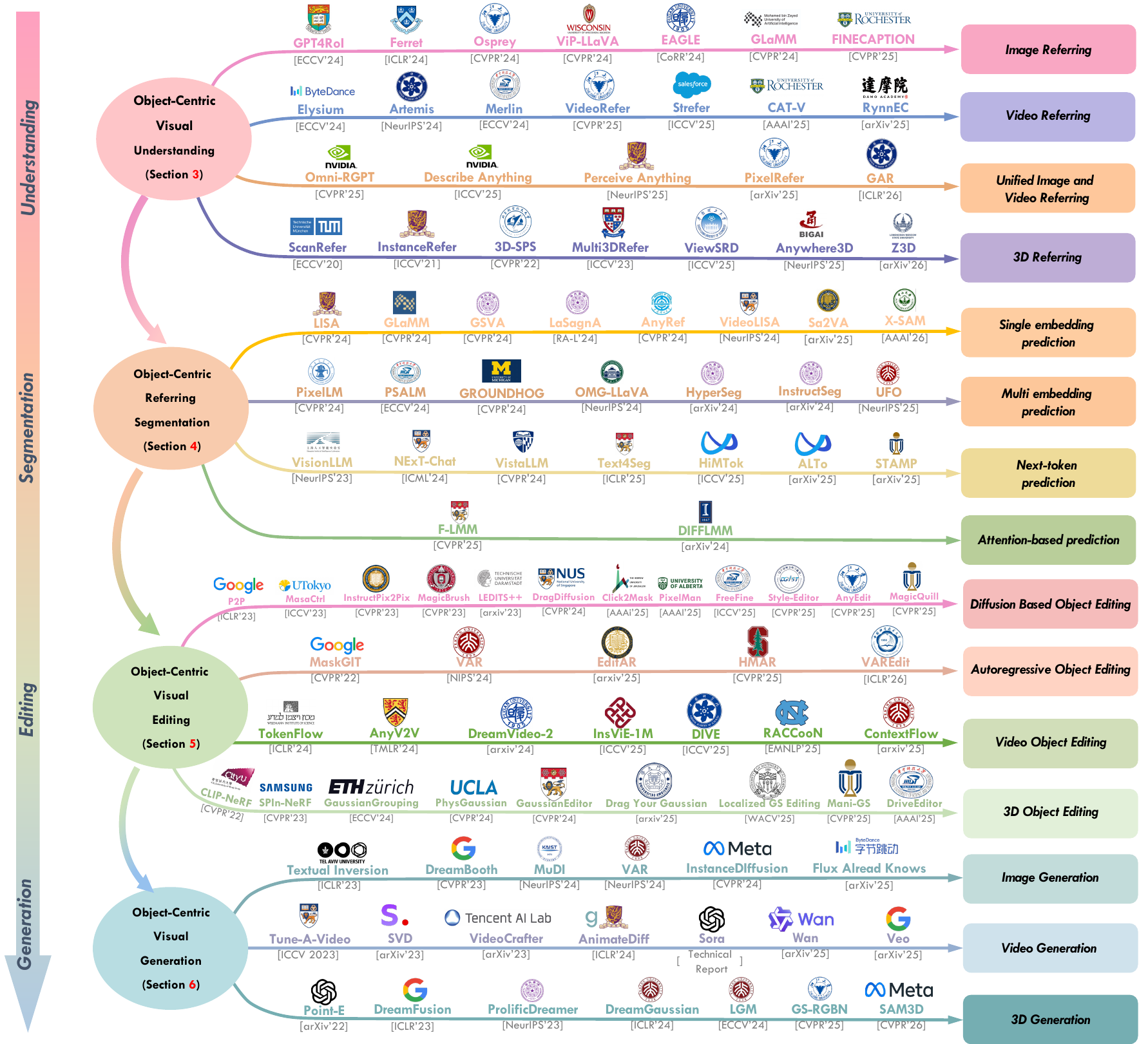}
\caption{
\textbf{Summary of representative models spanning object-centric visual understanding, segmentation, editing, and generation.} For a comprehensive list of related approaches, along with detailed discussions of their specifications, configurations, and technical aspects, please refer to Section~\ref{sec3:understanding}, Section~\ref{sec4:segmentation}, Section~\ref{sec5:editing}, and Section~\ref{sec6:generation}, respectively.
}
\label{fig:methods}
\end{figure*}










\section{Background and Preliminaries}
\label{sec2:background}
\subsection{Problem Definition}
To establish a unified framework for object-centric vision, we first define the fundamental notations used throughout this paper. Let $I \in \mathbb{R}^{H \times W \times 3}$ denote an input visual signal (e.g., an image or a video frame). Let $T = \{t_1, t_2, \dots, t_N\}$ represent a textual sequence, such as a user query or a referring expression. Let $S$ denote a spatial prompt, which can take various forms including coordinates, bounding boxes, or masks. The generalized objective of an object-centric multimodal system is to learn a mapping function $f_\theta: (I, T, S) \rightarrow O$, where $O$ is the task-specific object-centric output.

\subsubsection{Object Understanding}
Object-centric understanding requires the model to not only globally describe a scene but to identify, reason about, and comprehend specific instances based on spatial or textual cues. Given an image $I$, a spatial prompt $S$ specifying the target region, and a textual query $T_{query}$, the objective is to generate a descriptive or reasoning-based textual response $O_{text}$. Formally:
$$O_{text} = f_{understand}(I, S, T_{query}).$$
The core challenge lies in effectively projecting continuous spatial conditions $S$ or region-specific visual features into the discrete language space, enabling the model to perform fine-grained localized reasoning while maintaining global contextual awareness without suffering from object hallucination.

\subsubsection{Object Segmentation}
Object-centric segmentation partitions an image into meaningful regions with pixel-level precision based on open-ended linguistic descriptions. Given an input image $I$ and a textual query $T_{query}$, the objective is to map these inputs to a binary mask $O_{mask} \in \{0, 1\}^{H \times W}$ that identifies the target instance. Formally:
$$O_{mask} = f_{segment}(I, T_{query}).$$
Modern architectures typically achieve this alignment by predicting a specialized embedding (e.g., a \texttt{[SEG]} token), which is subsequently decoded using dense spatial features to bridge the inherent semantic gap between the language model's high-level reasoning and the vision model's fine-grained pixel-level priors.

\subsubsection{Object Editing}
Object editing involves the controlled manipulation of visual content while preserving desired attributes and non-target regions. Given a source image $I$, an editing instruction $T_{query}$, and an optional spatial condition $S$ (e.g., a user-provided mask or an implicitly extracted cross-attention map), the goal is to synthesize a modified image $O_{img}$. The problem is formulated as:
$$O_{img} = f_{edit}(I, S, T_{query}).$$
The core challenge lies in navigating the delicate trade-off between semantic flexibility and spatial fidelity. It requires the precise disentanglement of visual features to robustly maintain the source object's identity and global background consistency, while simultaneously executing highly localized, pixel-precise modifications that seamlessly blend with the unedited context without introducing unintended artifacts.

\subsubsection{Object Generation}
Object-centric generation synthesizes visual content conditioned on multimodal inputs under strict object-level constraints. Unlike global text-to-image synthesis, this task requires precise control over instance placement and attributes. 

Let $I_{\emptyset}$ denote the initial latent noise or empty canvas, $T_{query}$ the textual query, and $S$ the spatial prompt (e.g., bounding boxes or layouts) that defines the structural constraints. The generation of the output image $O_{img}$ is formulated as:
    $$O_{img} = f_{generate}(I_{\emptyset}, S, T_{query}).$$
The core challenge in this formulation transcends global text-to-image mapping. It demands complex, compositional cross-modal alignment. The model must strictly bind fine-grained textual attributes to their corresponding spatial regions explicitly designated by $S$, ensuring accurate geometric layouts while effectively mitigating severe issues like attribute leakage or identity confusion when synthesizing multiple interacting instances.

\subsection{Object Representation}
To bridge language-level reasoning and pixel-level execution, multimodal systems employ diverse object representations. The choice of representation dictates the precision and flexibility of the model:
\begin{itemize}
    \item \textbf{Bounding Boxes:} Represented by discrete coordinate tuples $(x_{min}, y_{min}, x_{max}, y_{max})$. They are easily integrated into language token sequences, but struggle to capture arbitrary-shaped regions or fine-grained boundaries.
    \item \textbf{Points:} Defined by $(x, y)$ coordinates, points provide a lightweight and interactive mechanism for spatial prompting, heavily utilized in dynamic mask generation and drag-based editing.
    \item \textbf{Masks:} Dense binary matrices $M \in \{0, 1\}^{H \times W}$ that offer precise pixel-level object boundaries. They are optimal for complex shapes and are widely used in robust referring segmentation pipelines.
    \item \textbf{Object Tokens:} High-dimensional embeddings $E \in \mathbb{R}^{N \times C}$ that encapsulate region-specific features, where $N$ denotes the number of object queries and $C$ represents the feature dimension. They enable the seamless integration of continuous visual descriptors with discrete linguistic tokens, supporting unified reasoning and cross-modal interaction within Transformer-based architectures.
    
\end{itemize}

\subsection{Basic Architecture}

\subsubsection{Multimodal Large Language Models}
Multimodal Large Language Models (MLLMs) serve as the cognitive backbone for object-centric tasks, evolving from complex modular alignment to streamlined end-to-end integration. Early foundational works explored diverse bridging mechanisms: \textbf{Flamingo}~\cite{alayrac2022flamingovisuallanguagemodel} pioneered \textbf{Gated Cross-Attention}, while \textbf{BLIP-2}~\cite{li2023blip2bootstrappinglanguageimagepretraining}, \textbf{InstructBLIP}~\cite{dai2023instructblipgeneralpurposevisionlanguagemodels}, and \textbf{MiniGPT-4}~\cite{zhu2023minigpt} introduced the \textbf{Q-Former}. More recently, the field has largely converged on coupling vision encoders with LLMs via lightweight projection layers~\cite{liu2023visual, li2025tokenpacker,zhang2024eagle,chen2024fargpt4vclosinggap}.
From \textbf{LLaVA-1.5}~\cite{liu2023visual} to dynamic-resolution \textbf{LLaVA-NeXT}~\cite{liu2024llavanext} and the unified image–video \textbf{LLaVA-OneVision}~\cite{li2024llavaonevisioneasyvisualtask}, this line of work continues to advance general-purpose multimodal understanding.
The open-weight community has further driven progress toward large-scale modeling and fine-grained spatial perception. The \textbf{InternVL} family~\cite{chen2024fargpt4vclosinggap, gao2024mini, zhu2025internvl3exploringadvancedtraining, wang2025internvl3} systematically explores scaling laws by expanding vision encoders to billions of parameters and introducing techniques such as Cascade Reinforcement Learning and the Visual Resolution Router (ViR) for long videos. In parallel, \textbf{Qwen-VL} iterations~\cite{wang2024qwen2vlenhancingvisionlanguagemodels, bai2025qwen25vltechnicalreport, bai2025qwen3} emphasize spatial fidelity through Native Dynamic Resolution and M-RoPE. Together, these advances shift models from basic captioning toward more capable, agentic behavior, including pixel-level grounding and long-form video understanding. While proprietary systems such as \textbf{GPT-5.2}~\cite{2023GPT4VisionSC}, \textbf{Gemini 3 Pro}, and \textbf{Claude Opus 4.5} remain industry benchmarks, open-weight models provide a strong foundation for specialized and extensible applications.

\subsubsection{Vision Foundation Models for Object Perception}
Although MLLMs are strong at high-level semantic reasoning, their flattened token representations often struggle with fine-grained spatial localization. Vision foundation models therefore play a complementary role by providing dense spatial priors for object-centric perception. Recent progress has also shifted from closed-set recognition to open-vocabulary grounding. \textbf{DINOv2}~\cite{oquab2024dinov2learningrobustvisual} improves self-supervised visual representation learning, while \textbf{Grounding DINO}~\cite{liu2024groundingdinomarryingdino} and \textbf{Grounding DINO 1.5}~\cite{ren2024groundingdino15advance} align region-level features with textual queries to support accurate open-vocabulary object localization in complex scenes. Beyond boxes, promptable segmentation has become a key component of dense perception. \textbf{SAM}~\cite{lin2025perceive} established a scalable zero-shot segmentation paradigm, alongside efficient variants developed to reduce computational cost~\cite{sun2025efficientvariantssegmentmodel}, and \textbf{SAM~2}~\cite{ravi2024sam2segmentimages} extended it to video through a unified spatio-temporal architecture with memory attention for consistent object tracking. More recent models such as \textbf{SAM~3}~\cite{carion2025sam3segmentconcepts} further improve high-resolution perception and cross-modal alignment, particularly for large-scale and egocentric video understanding. At the same time, perception is increasingly moving toward unified decoding frameworks. Models such as \textbf{X-Decoder}~\cite{zou2022generalizeddecodingpixelimage} and \textbf{SEEM}~\cite{zou2023segment} provide a common interface for segmentation, grounding, and localized captioning, making dense visual perception easier to integrate into MLLMs.

\subsubsection{Generative Models}
Generative models act as the creative engines of object-centric pipelines, complementing perception models by enabling localized editing, inpainting, and high-fidelity synthesis for downstream tasks. The field is undergoing a paradigm shift from traditional U-Net architectures to scalable transformer-based designs, leading to substantial gains in structural fidelity and spatiotemporal consistency. Building on latent diffusion foundations such as \textbf{Stable Diffusion} (\textbf{SD1.5}~\cite{rombach2022high}, \textbf{SDXL}~\cite{podell2023sdxlimprovinglatentdiffusion}), recent advances address the longstanding challenge of composing multiple objects with precise spatial relationships. \textbf{Stable Diffusion 3 (SD3)}~\cite{esser2024scalingrectifiedflowtransformers} introduces rectified flow transformers to improve compositional accuracy, while flow-based models like \textbf{Flux.1}~\cite{greenberg2025demystifyingfluxarchitecture} further enhance text-to-image alignment and structural coherence, supporting accurate object generation, text rendering, and fine-grained spatial control. Extending generation to the temporal domain, Diffusion Transformers (DiT) have enabled a shift toward coherent physical world simulation, as demonstrated by models such as OpenAI’s \textbf{Sora}~\cite{liu2024sorareviewbackgroundtechnology} and Google’s \textbf{Veo}, which produce long-duration videos with consistent object identity, texture, and interaction under complex motion and occlusion. The latest progress focuses on fine-grained controllability and multimodal synchronization, where region-level editing via masking and latent interpolation allows precise manipulation of individual objects without disrupting the background, and systems like \textbf{Kling 2.6}~\cite{klingteam2025klingomnitechnicalreport} and \textbf{Sora 2} further advance native audio-visual alignment and timeline-based control, enabling detailed spatiotemporal editing and dynamic object-level animation.
\section{Object-Centric Visual Understanding}
\label{sec3:understanding}

\subsection{Architecture}
Unlike general vision-language tasks, object-centric understanding necessitates the construction of precise region-level representations. Current approaches mainly evolve across two dimensions: the paradigm for introducing object prompts, and the mechanism for encoding target regions.

\subsubsection{Paradigm}


Existing paradigms mainly differ in how object cues are incorporated into MLLMs, ranging from textualized prompts to explicit visual tokens and feature-level fusion. This progression reflects a shift from language-side prompting to more grounded object modeling.

\begin{figure*}[t]
  \centering
  \includegraphics[width=0.9\textwidth]{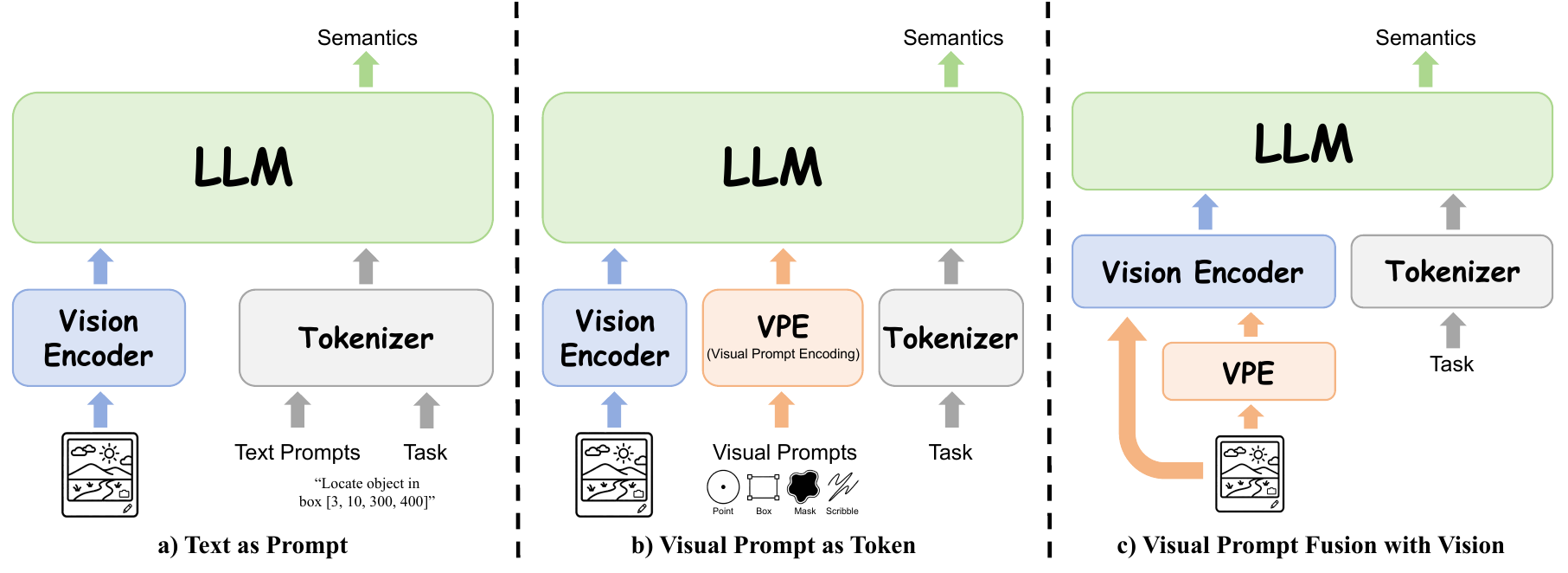} 
  \caption{Illustration of three representative paradigms for incorporating visual prompts into multimodal large language models: Text as Prompt, Visual Prompt as Token, and Visual Prompt Fusion with Vision. These paradigms differ in whether visual prompts are expressed as language, injected as prompt tokens, or fused directly with visual representations before language modeling.}
  \label{fig:understanding}
\end{figure*}

\paragraph{Text as Prompt}
A straightforward paradigm is to encode object cues as textualized spatial prompts and feed them into MLLMs through the language interface. \textbf{Kosmos2}~\cite{peng2023kosmos} links text spans to image regions via location tokens, \textbf{Shikra}~\cite{chen2023shikra} treats spatial coordinates as natural-language inputs and outputs, \textbf{Pink}~\cite{xuan2024pink} represents region cues with textualized box coordinates, and \textbf{ChatSpot}~\cite{zhao2023chatspot} extends this paradigm to more precise referring instruction tuning. Its advantage is simplicity, as existing MLLM architectures can be largely reused. However, because region priors are mainly injected on the language side, this paradigm is less effective for arbitrary-shaped regions, fine-grained boundaries, and heavy occlusion. It has also been extended to videos by \textbf{Elysium}~\cite{wang2024elysium}, although maintaining object identity over time remains challenging.

\paragraph{Visual Prompt as Token}
A natural extension of text-based prompting is to encode object cues as tokenized or marker-based visual prompts and feed them jointly with image and text tokens into MLLMs. Compared with text-as-prompt methods, this paradigm provides more direct region-language alignment. \textbf{Draw-and-Understand}~\cite{lin2024draw} encodes points, boxes, and free-form shapes with a visual prompt encoder, while \textbf{Ferret}~\cite{you2023ferret} employs a hybrid region representation that integrates coordinate-based cues with continuous visual embeddings. \textbf{Ferret-v2}~\cite{zhang2024ferret} further strengthens this line with any-resolution grounding and multi-granularity visual encoding. \textbf{Osprey}~\cite{Osprey} further injects mask regions as visual and spatial tokens, and \textbf{RegionGPT}~\cite{guo2024regiongpt} replaces region placeholders with semantic region embeddings. Related visual-prompt variants such as \textbf{ViP-LLaVA}~\cite{cai2024vip} and \textbf{EAGLE}~\cite{zhang2024eagle} directly preserve arbitrary visual markers on the image canvas, enabling prompt comprehension without heavy region encoders. Extending this idea to unified image and video understanding, \textbf{Omni-RGPT}~\cite{heo2025omni} introduces Token Mark for more consistent region representation across modalities and time. Overall, this paradigm enables more direct region-level interaction, although its effectiveness still depends on the design and alignment of prompt representations.

\paragraph{Visual Prompt Fusion with Vision}
A major direction is to explicitly extract region features and fuse them with global visual representations for LLM reasoning. Early efforts focus on feature-level region injection. \textbf{GPT4RoI}~\cite{zhang2025gpt4roi} replaces region references with RoI features interleaved with the instruction sequence. Related efforts such as \textbf{Alpha-CLIP}~\cite{sun2024alpha} and \textbf{PerceptionGPT}~\cite{pi2024perceptiongpt} further enhance region-sensitive visual perception for multimodal reasoning. Subsequent works move toward pixel and mask-aware modeling. \textbf{GLaMM}~\cite{GLaMM} connects region understanding with grounded mask generation, while \textbf{DAM}~\cite{lian2025describe} and \textbf{GAR}~\cite{wang2025grasp} emphasize localized understanding and context preservation. \textbf{PAM}~\cite{lin2025perceive} further adopts SAM~2 as the visual backbone and converts segmentation features into multimodal tokens, while \textbf{PixelRefer}~\cite{yuan2025pixelrefer} extends this paradigm to unified image-video object understanding with compact object tokens and early context infusion. This paradigm yields more precise, grounded object representations, but requires careful design for feature fusion, context preservation, and efficiency.

\subsubsection{Region Encoder}
From another perspective, existing methods can be categorized by the form of region input. Different region encoders transform these cues into object-aware representations, affecting region precision and flexibility in image and video understanding. Current approaches mainly take coordinates or bounding boxes, masks or free-form regions, and more diverse visual prompts as input.

\paragraph{Coordinates and Bounding Boxes as Input}
A basic input form is to specify the target region with coordinates or bounding boxes. \textbf{Kosmos2}~\cite{peng2023kosmos} represents grounded regions with location tokens, \textbf{Shikra}~\cite{chen2023shikra} treats spatial coordinates as natural-language inputs and outputs, and \textbf{Pink}~\cite{xuan2024pink} encodes bounding box coordinates in textual form for referential comprehension. Moving beyond purely textualized inputs, \textbf{GPT4RoI}~\cite{zhang2025gpt4roi} injects box-conditioned RoI features into the instruction sequence, enabling region-level interaction and reasoning. Overall, this input form is simple and effective, but it mainly captures coarse spatial extent and is less suitable for arbitrary-shaped regions or fine-grained boundaries.

\paragraph{Masks and Free-form Regions as Input}
Another input form is to specify the target with masks or free-form regions, which better preserve arbitrary shapes and fine-grained boundaries. \textbf{Osprey}~\cite{Osprey} supports mask-based region referring for pixel-level understanding. \textbf{FINECAPTION}~\cite{hua2025finecaption} further explores arbitrary masks for fine-grained region description. Extending this idea to videos, \textbf{VideoRefer Suite}~\cite{yuan2025videorefer} adopts a unified pixel-level mask representation for single-frame and multi-frame free-form region inputs. \textbf{PixelRefer}~\cite{yuan2025pixelrefer} further generates compact object representations from free-form regions for unified image--video referring. Compared with coordinates or boxes, this input form is more suitable for complex shapes and pixel-level grounding.

\paragraph{Multi-format Visual Prompts as Input}
A more flexible input form is to support diverse visual prompts, such as points, boxes, masks, scribbles, or free-form shapes. \textbf{Draw-and-Understand}~\cite{lin2024draw}, \textbf{Ferret}~\cite{you2023ferret}, and \textbf{Ferret-v2}~\cite{zhang2024ferret} explicitly support arbitrary-shape or multi-granularity region prompts. Related variants such as \textbf{ViP-LLaVA}~\cite{cai2024vip} and \textbf{EAGLE}~\cite{zhang2024eagle} directly preserve arbitrary visual markers on the image canvas, while \textbf{DAM}~\cite{lian2025describe} further unifies clicks, scribbles, boxes, and masks for localized understanding. Compared with single-format inputs, this design offers greater flexibility for interactive region understanding.

\subsection{Image Referring} 

Image referring studies how MLLMs interpret user-specified objects or regions in an image, given either a referring expression or explicit spatial prompts such as boxes, points, or masks, to enable open-ended region-level understanding, including recognition, description, and reasoning over the referred content.

\subsubsection{Region Cognition}
A central line of image referring is to move from object localization to region cognition, namely, recognizing and understanding what the referred region contains. \textbf{GPT4RoI}~\cite{zhang2025gpt4roi} enables region-level interaction and reasoning through RoI-conditioned instruction tuning, while \textbf{Ferret}~\cite{you2023ferret} extends this capability to free-form and multi-granularity regions. \textbf{Osprey}~\cite{Osprey} further pushes region cognition to pixel-level understanding with mask-based inputs. Related efforts such as \textbf{Pink}~\cite{xuan2024pink}, \textbf{ViP-LLaVA}~\cite{cai2024vip}, \textbf{EAGLE}~\cite{zhang2024eagle}, and \textbf{ChatSpot}~\cite{zhao2023chatspot} further strengthen referential comprehension under textual, visual-marker, and interactive prompts. Together, these works mark the shift from grounding where an object is to understanding what a referred region is.

\subsubsection{Region Detailed Description}
Beyond recognizing a referred region, recent work aims to describe it in detail. \textbf{DAM}~\cite{lian2025describe} formulates this setting as detailed localized captioning, emphasizing rich region descriptions with both local detail and global context. \textbf{FINECAPTION}~\cite{hua2025finecaption} further supports arbitrary masks as referential inputs for multi-grained compositional region captioning. Related efforts such as \textbf{Caption Anything}~\cite{wang2023caption} also highlight controllable region description under interactive visual prompts. Together, these works extend image referring from region recognition toward fine-grained and descriptive region understanding.

\subsubsection{Region-level QA and Dialogue}
Recent work further extends image referring to question answering and grounded dialogue. \textbf{GPT4RoI}~\cite{zhang2025gpt4roi} enables region-conditioned reasoning and supports question answering over referred regions through RoI-guided instruction tuning. \textbf{GLaMM}~\cite{GLaMM} further connects region understanding with grounded conversation generation by coupling language responses with object masks.

\subsubsection{Contextual Reasoning}
Image referring further extends to contextual reasoning, where understanding a referred region requires integrating surrounding objects, events, and commonsense context. \textbf{VCR}~\cite{zellers2019recognition} serves as a representative benchmark for this direction by requiring models to answer visual questions and justify them with rationales. Building on this line, \textbf{GAR}~\cite{wang2025grasp} improves contextual and compositional reasoning through RoI-aligned feature replay, while \textbf{RCMU/RCVIT}~\cite{wei2025region} further emphasizes incorporating object-related textual context into region-level understanding.

\subsubsection{Spatial Reasoning}
Another direction is spatial reasoning over referred regions. \textbf{SpatialRGPT}~\cite{cheng2024spatialrgpt} improves the spatial understanding capabilities of vision-language models, enabling reasoning about directions, distances, and spatial arrangements from user-specified region proposals. This line shows that image referring also requires grounded reasoning over spatial structure.

\subsection{Video Referring}


Video referring extends MLLM understanding to specific objects or regions across frames via text or spatial prompts. Unlike static image grounding, it requires consistent spatio-temporal tracking and interpretation, framing it as a joint spatio-temporal understanding challenge.

\subsubsection{Temporal Region Cognition}
A central line of video referring is temporal region cognition, namely, recognizing and understanding a referred object or region across time. \textbf{Elysium}~\cite{wang2024elysium} formulates object-level perception in videos, \textbf{Artemis}~\cite{qiu2024artemis} studies referential understanding from a question and a box in complex videos, and \textbf{Merlin}~\cite{han2024merlin} further broadens this direction with object-centric video understanding and future-aware reasoning.

\subsubsection{Temporally Detailed Region Description}
Recent work further extends video referring to temporally detailed region description. \textbf{DAM}~\cite{lian2025describe} formulates this setting as detailed localized captioning for user-specified regions in images and videos, while \textbf{PAM}~\cite{lin2025perceive}, \textbf{CAT-V}~\cite{tang2025caption}, \textbf{Omni-RGPT}~\cite{heo2025omni}, and \textbf{PixelRefer}~\cite{yuan2025pixelrefer} further support fine-grained region-specific semantic outputs or object-centric video descriptions.

\subsubsection{Region-level Video QA and Dialogue}
Video referring also extends to region-level question answering and dialogue over time. \textbf{VideoRefer Suite}~\cite{yuan2025videorefer} supports perception and reasoning on user-specified objects throughout a video, while \textbf{PixelRefer}~\cite{yuan2025pixelrefer} introduces multi-turn QA over referred regions. \textbf{Omni-RGPT}~\cite{heo2025omni} further strengthens region-level video interaction with more consistent object representations.

\subsubsection{Dynamic Contextual Reasoning}
Beyond understanding a target in isolation, video referring increasingly requires reasoning in dynamic context. \textbf{GAR}~\cite{wang2025grasp} provides an image-side precursor for context-aware region reasoning, while in videos \textbf{Omni-RGPT}~\cite{heo2025omni} supports context-aware region comprehension across time and \textbf{Strefer}~\cite{zhou2025strefer} explicitly emphasizes space-time referring and reasoning in dynamic scenes.

\subsubsection{Embodied Spatial-temporal Reasoning}
Another emerging direction is embodied spatial-temporal reasoning over referred regions. \textbf{RynnEC}~\cite{dang2025rynnec} enhances embodied cognition and spatial reasoning in video MLLMs, while \textbf{SR-3D}~\cite{cheng20253d} supports 3D-aware region prompting and cross-frame spatial reasoning through a shared 2D/3D visual token space.

\subsection{Unified Image and Video Referring} Unified image and video referring studies how MLLMs understand an object or region across images and videos under a shared interface. 
Recent works begin to formulate image and video referring within unified frameworks. \textbf{Omni-RGPT}~\cite{heo2025omni} emphasizes shared region representation through Token Mark and improves temporal consistency in video understanding. \textbf{DAM}~\cite{lian2025describe} highlights a unified prompting interface, supporting points, boxes, scribbles, and masks for localized image and video captioning. \textbf{PAM}~\cite{lin2025perceive} further extends unified referring to broader region-level semantic understanding by combining segmentation with recognition, explanation, and captioning. \textbf{PixelRefer}~\cite{yuan2025pixelrefer} advances unified spatio-temporal object referring with arbitrary granularity through scale-adaptive object representations.

\subsection{3D Referring}
3D referring studies how a model localizes, identifies, or a 3D target from a natural-language expression. Unlike 2D referring, it grounds language in structured 3D observations, such as point clouds, RGB-D scans, or multi-view reconstructions. The core challenge is to align language with 3D geometry, spatial layout and object relations.

\subsubsection{Foundations}
Foundational 3D referring studies were established by \textbf{ScanRefer}~\cite{chen2020scanrefer} and \textbf{ReferIt3D}~\cite{achlioptas2020referit3d}. \textbf{ScanRefer} formulates natural language grounding in RGB-D scans from free-form descriptions, while \textbf{ReferIt3D} investigates fine-grained object recognition in real-world 3D environments and introduces the \textbf{Nr3D} and \textbf{Sr3D} benchmarks, highlighting instance-level distinction and spatially explicit referring expressions. Together, these works establish the basic task setting and benchmark foundation for subsequent research.

\subsubsection{Proposal-based Understanding}
Early 3D referring methods largely relied on proposal- or instance-centric pipelines to support target understanding in complex 3D scenes. \textbf{InstanceRefer}~\cite{yuan2021instancerefer} formulates 3D referring as instance matching by integrating candidate filtering, instance attributes, inter-instance relations, and localization cues. \textbf{3DVG-Transformer}~\cite{zhao20213dvg} further strengthens this paradigm with transformer-based proposal modeling and cross-modal interaction, improving target identification and disambiguation. Moving beyond two-stage designs, \textbf{3D-SPS}~\cite{luo20223d} introduces a single-stage framework that progressively selects language-relevant points to localize the target more directly. Together, these works establish proposal-based target understanding as a major paradigm in early 3D referring.

\subsubsection{Relation Reasoning}
Spatial relation modeling is central to 3D referring, since many expressions depend more on relative position than object appearance. \textbf{ViL3DRel}~\cite{chen2022language} places language-conditioned spatial reasoning at the core of 3D grounding through pairwise distances and orientations. \textbf{3DRP-Net}~\cite{wang20233drp} further emphasizes relative position-aware modeling in a one-stage framework. \textbf{ViewSRD}~\cite{huang2025viewsrd} extends this direction to recent multi-anchor queries with structured multi-view decomposition. Together, these works establish relation reasoning as a key direction in 3D referring.

\subsubsection{Generalized and Unified}
Recent 3D referring has moved beyond single-object setting toward generalized and unified formulations. \textbf{Multi3DRefer}~\cite{zhang2023multi3drefer} extends 3D grounding to zero-, single-, and multiple-object targets, while \textbf{Uni3DL}~\cite{li2024uni3dl} places visual grounding in a unified 3D vision-language framework. \textbf{3D-GRAND}~\cite{yang2024_3D_GRAND} scales with million-scale grounded 3D-text data, and \textbf{Anywhere3D-Bench}~\cite{anywhere3d} broadens grounding beyond objects to multi-level targets such as activity areas, free space, and object parts. \textbf{Z3D}~\cite{drozdov2026z3d} further extends this direction toward zero-shot transfer. Together, these works show that 3D referring is evolving toward broader, unified, and open-ended 3D grounding.

\subsection{Datasets and Benchmarks}

The development of object-centric visual understanding is supported by diverse datasets and benchmarks spanning region grounding, referring expression comprehension, localized captioning, object tracking, and multimodal reasoning. As shown in Table~\ref{tab:ocvu}, early image datasets such as \textbf{LVIS}~\cite{gupta2019lvis}, \textbf{Visual Genome}~\cite{krishna2017visual}, \textbf{PACO}~\cite{ramanathan2023paco}, and \textbf{Flickr30k Entities}~\cite{plummer2015flickr30k} provide important foundations for instance-level, part-level, and phrase-region aligned understanding. More recent instruction-tuning datasets, including \textbf{GRIT}~\cite{you2023ferret}, \textbf{PAM-Data}~\cite{lin2025perceive}, \textbf{Osprey-724K}~\cite{Osprey}, and \textbf{PixelRefer-2.2M}~\cite{yuan2025pixelrefer}, further extend this line toward fine-grained region-level and pixel-level multimodal understanding.

In the video domain, datasets such as \textbf{HC-STVG}~\cite{tang2021human}, \textbf{Vid-STG}~\cite{zhang2020does}, \textbf{ElysiumTrack-1M}~\cite{wang2024elysium}, \textbf{RegVID-300K}~\cite{heo2025omni}, and \textbf{VideoRefer-700K}~\cite{yuan2025videorefer} promote progress from spatio-temporal grounding to object-level tracking, region-level captioning, and video referring. Benchmarks such as \textbf{Ferret-Bench}~\cite{you2023ferret}, \textbf{Ref-L4}~\cite{chen2025revisiting}, and \textbf{BenSMOT}~\cite{li2024beyond} further evaluate grounding, referring, and temporally consistent object understanding under more challenging settings.
Overall, current datasets and benchmarks show a clear trend from static object recognition toward localized, interactive, and temporally consistent object understanding across both images and videos.

\begin{center}
    \begin{table*}[t]

\centering
\captionsetup{justification=raggedright, singlelinecheck=false}
\caption{Summary of representative \textbf{Datasets \& Benchmarks} for \textbf{object-centric visual understanding}.
    \\
    $\bullet$ \textbf{Modality}: \iconImg: Image, \iconVid: Video.
    \\
    $\bullet$ \textbf{Role}: \bdgDataset: Training Dataset, and \bdgBenchmark: Evaluation Benchmark.
}
\label{tab:ocvu}
\vspace{-0.2cm}
\resizebox{\linewidth}{!}{
\begin{tabular}{llcccl}
    \toprule
    \textbf{Dataset \& Benchmark} & \textbf{Year} & \textbf{Modality} \raisebox{-0.25ex}{\includegraphics[width=0.022\linewidth]{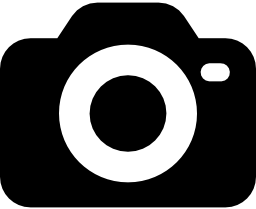}} & \textbf{Scale} & \textbf{Role} & \textbf{Task Focus} \raisebox{-0.5ex}{\includegraphics[width=0.021\linewidth]{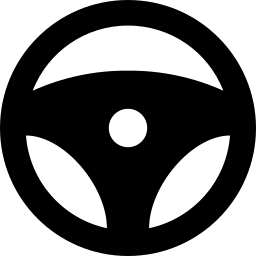}}
    \\
    \midrule
    \midrule

    GRIT \cite{you2023ferret}
    & 2024
    & \iconImg
    & 1.1M
    & \bdgDataset
    & Region grounding and referring
    \\
    \rowcolor{gray!7}
    PAM-Data \cite{lin2025perceive}
    & 2025
    & \iconImg \iconVid
    & 2.1M
    & \bdgDataset
    & Region recognition and explanation
    \\
    LVIS \cite{gupta2019lvis}
    & 2019
    & \iconImg
    & 164K
    & \bdgDataset
    & Large-vocabulary object instances
    \\
    \rowcolor{gray!7}
    Visual Genome \cite{krishna2017visual}
    & 2017
    & \iconImg
    & 108K
    & \bdgDataset
    & Objects, regions, and relations
    \\
    HC-STVG \cite{tang2021human}
    & 2020
    & \iconVid
    & 5.6K
    & \bdgDataset
    & Human-centric spatio-temporal grounding
    \\
    \rowcolor{gray!7}
    ElysiumTrack-1M \cite{wang2024elysium}
    & 2024
    & \iconVid
    & 1.27M
    & \bdgDataset
    & Object-level video tracking
    \\
    Describe Anything \cite{lian2025describe}
    & 2025
    & \iconImg \iconVid
    & 916K
    & \bdgDataset
    & Localized image/video description
    \\
    \rowcolor{gray!7}
    PACO \cite{ramanathan2023paco}
    & 2023
    & \iconImg \iconVid
    & 84K
    & \bdgDataset
    & Part-level object understanding
    \\
    Flickr30k Entities \cite{plummer2015flickr30k}
    & 2015
    & \iconImg
    & 30K
    & \bdgDataset
    & Phrase grounding
    \\
    \rowcolor{gray!7}
    RegVID-300K \cite{heo2025omni}
    & 2025
    & \iconVid
    & 300K
    & \bdgDataset
    & Video regions and referring
    \\
    Vid-STG \cite{zhang2020does}
    & 2020
    & \iconVid
    & 10K
    & \bdgDataset
    & Spatio-temporal video grounding
    \\
    \rowcolor{gray!7}
    VideoRefer-700K \cite{yuan2025videorefer}
    & 2025
    & \iconVid
    & 700K
    & \bdgDataset
    & Object-level video instruction tuning
    \\
    MDVP \cite{lin2024draw}
    & 2024
    & \iconImg
    & 1.2M
    & \bdgDataset
    & Multi-dimensional visual perception
    \\
    \rowcolor{gray!7}
    RefCOCO/RefCOCO+/RefCOCOg \cite{mao2016generation}
    & 2016
    & \iconImg
    & 142K
    & \bdgDataset
    & Referring expression grounding
    \\
    Osprey-724K \cite{Osprey}
    & 2024
    & \iconImg
    & 724K
    & \bdgDataset
    & Mask-based pixel understanding
    \\
    \rowcolor{gray!7}
    LVIS \cite{gupta2019lvis}
    & 2019
    & \iconImg
    & 5K
    & \bdgBenchmark
    & Large-vocabulary object understanding
    \\
    Visual Genome \cite{krishna2017visual}
    & 2017
    & \iconImg
    & 5.8K
    & \bdgBenchmark
    & Scene graph and relation grounding
    \\
    \rowcolor{gray!7}
    HC-STVG \cite{tang2021human}
    & 2020
    & \iconVid
    & 1.16K
    & \bdgBenchmark
    & Human-centric spatio-temporal grounding
    \\
    PACO \cite{ramanathan2023paco}
    & 2023
    & \iconImg \iconVid
    & 9.4K
    & \bdgBenchmark
    & Part-level object understanding
    \\
    \rowcolor{gray!7}
    Flickr30k Entities \cite{plummer2015flickr30k}
    & 2015
    & \iconImg
    & 1K
    & \bdgBenchmark
    & Phrase grounding
    \\
    VideoRefer-Bench-D\cite{yuan2025videorefer}
    & 2025
    & \iconVid
    & 400
    & \bdgBenchmark
    & Object-level video description
    \\
    VideoRefer-Bench-Q\cite{yuan2025videorefer}
    & 2025
    & \iconVid
    & 1000
    & \bdgBenchmark
    & Object-level video QA
    \\
    \rowcolor{gray!7}
    Ferret-Bench \cite{you2023ferret}
    & 2024
    & \iconImg
    & 120
    & \bdgBenchmark
    & Referring, grounding, and reasoning
    \\
    Ref-L4 \cite{chen2025revisiting}
    & 2024
    & \iconImg
    & 45K
    & \bdgBenchmark
    & Fine-grained localized reasoning
    \\
    \rowcolor{gray!7}
    Vid-STG \cite{zhang2020does}
    & 2020
    & \iconVid
    & 10K
    & \bdgBenchmark
    & Spatio-temporal video grounding
    \\
    BenSMOT \cite{li2024beyond}
    & 2024
    & \iconVid
    & 3.3K
    & \bdgBenchmark
    & Semantic multi-object tracking
    \\

    \bottomrule
\end{tabular}}
\label{tab:object_centric_all}
\end{table*}
\end{center}

\section{Object-Centric Referring Segmentation}
\label{sec4:segmentation}
Object-centric referring segmentation extends object-centric understanding to precise delineation. Given linguistic or multimodal references, the goal is to identify the intended object and segment it at the pixel level. In the large-model era, this setting has evolved from category-closed supervision toward instruction following, ambiguity resolution, and reasoning-intensive mask prediction. In LMM-based systems, segmentation is therefore no longer only a dense prediction problem~\cite{wang2020solov2,li2022box}, but also a language-conditioned object grounding problem. We first examine the architectural design space and then review representative image, video, audio-visual, and emerging 3D settings.

\subsection{Architecture}
\noindent The core architectural problem in LMM-based referring segmentation is how to build a reliable language--pixel interface, namely, how to convert a free-form instruction and the model's reasoning into a spatial signal that can be executed as a mask. Existing systems can be grouped into four recurring patterns: \textbf{single embedding prediction}, \textbf{multi embedding prediction}, \textbf{next-token prediction}, and \textbf{attention-based prediction}. The first two produce compact prompts for a mask head, differing in whether the model emits one or multiple mask-aware embeddings; the third generates the mask itself or a structured proxy as a sequence; and the fourth derives the spatial signal from attention before mask refinement. This decomposition is compatible with broader  abstractions~\cite{MRSegSurvey} while remaining close to how recent papers actually instantiate the language--pixel interface.

\begin{figure*}[t]
\centering
\includegraphics[width=0.96\linewidth]{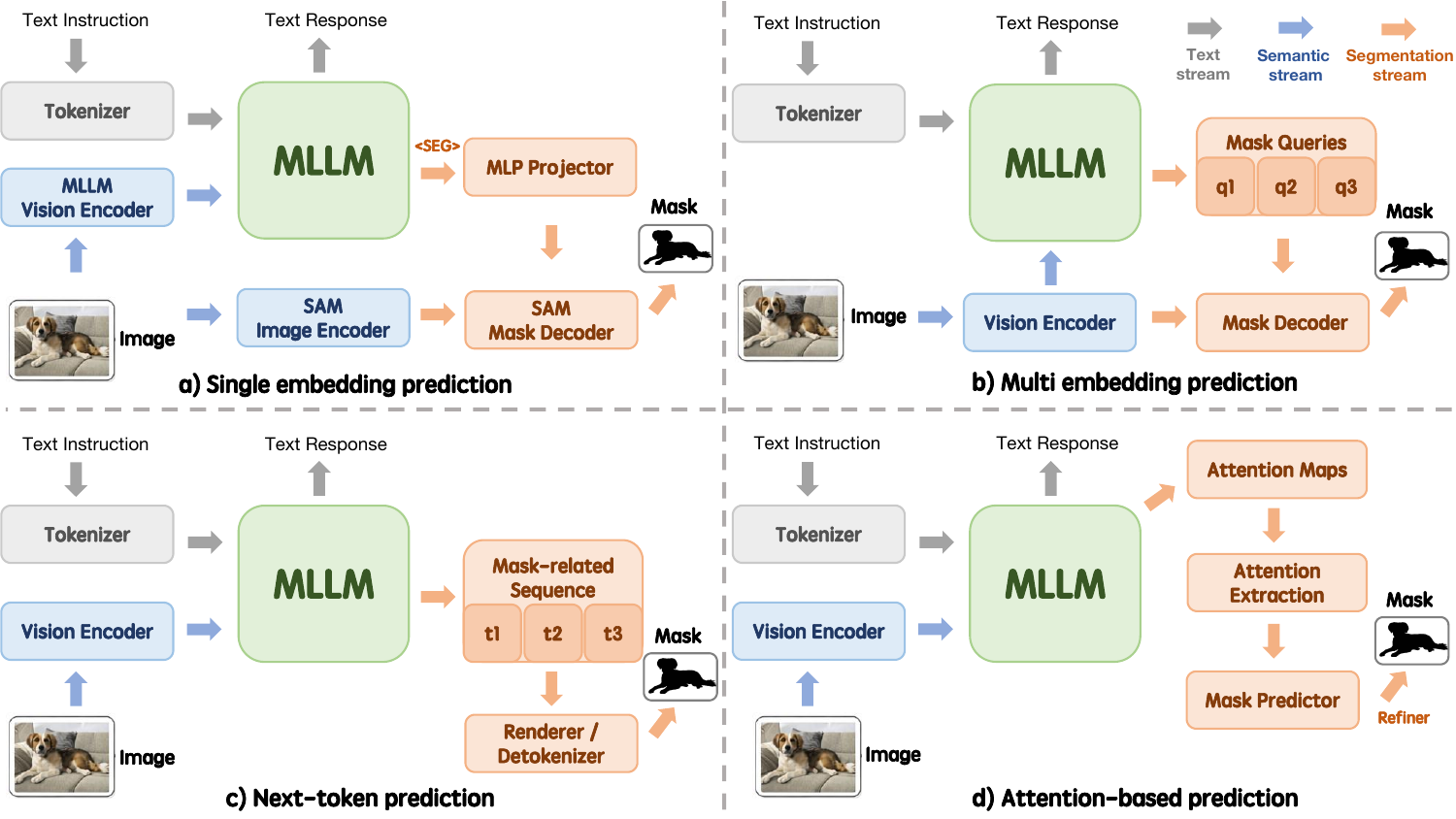}
\caption{Four representative language--pixel interfaces for LMM-based object-centric referring segmentation: (a) single embedding prediction, (b) multi embedding prediction, (c) next-token prediction, and (d) attention-based prediction. They differ in whether the segmentation signal is expressed as one prompt embedding, multiple mask-aware queries, a generated mask-related sequence, or an attention-derived spatial field.}
\label{fig:segmentation-architecture}
\end{figure*}

\subsubsection{Language--Pixel Interface}

\paragraph{Single embedding prediction.} In this family, the MLLM first resolves the instruction into one special segmentation-aware hidden state, and a downstream promptable segmenter then converts that compact embedding into a final mask. \textbf{LISA}~\cite{LISA} is a representative example. Given an image and text query, the MLLM generates  output sequence contains a special token such as \texttt{<SEG>}. The last-layer hidden state of this token is projected by a small MLP into the prompt space of a segmentation decoder; the decoder then combines the projected token with those dense features to produce the mask~\cite{LISA}. \textbf{GSVA}~\cite{GSVA} keeps the same prompt-to-mask pipeline, but predicts multiple \texttt{[SEG]} tokens and filters out \texttt{[REJ]} tokens so that multiple referred instances can be segmented or absent ones rejected~\cite{GSVA}. \textbf{GLaMM}~\cite{GLaMM}, \textbf{AnyRef}~\cite{AnyRef}, \textbf{VideoLISA}~\cite{VideoLISA}, and \textbf{Sa2VA}~\cite{Sa2VA} largely preserve this recipe while broadening the prompt source or extending the final executor from image segmentation to video mask propagation. The mask is not produced directly by the MLLM; it outputs one or a few prompt embeddings, and a downstream segmenter turns them into pixels.

\paragraph{Multi embedding prediction.} In this family, the instruction is translated into several segmentation-relevant embeddings or queries, and the final mask is obtained by jointly decoding them with dense visual features inside the model. The crucial difference from the previous family is that the target is not compressed into one prompt vector. In \textbf{PixelLM}~\cite{PixelLM}, a segmentation codebook is inserted into the LLM input together with the image and text; after autoregressive decoding, the hidden states corresponding to the codebook tokens are collected scale by scale and fed, together with projected image features, into a lightweight pixel decoder that outputs the mask~\cite{PixelLM}. \textbf{PSALM}~\cite{PSALM} follows the same overall logic with a unified schema of image, task instruction, conditional prompts, and mask tokens, and then uses a dedicated mask decoder to generate and classify masks~\cite{PSALM}. \textbf{GROUNDHOG}~\cite{GROUNDHOG} makes the query-centric design explicit by forming visual entity tokens from masked visual features and retrieving or merging the corresponding entity masks, while \textbf{OMG-LLaVA}~\cite{OMGLLaVA} couples injected perception priors and visual prompts with a universal segmentation decoder in an end-to-end design~\cite{OMGLLaVA}. The segmentation result is obtained by jointly decoding several mask-related latent variables, rather than by asking one token to summarize the entire target.

\paragraph{Next-token prediction.} In this family, segmentation is treated as sequence generation: the MLLM directly emits the mask itself or a structured proxy, which is then interpreted or decoded into the final pixel map. \textbf{VisionLLM}~\cite{VisionLLM} is an early example showing that segmentation outputs can be folded into the same open-ended generation interface as other vision tasks. \textbf{NExT-Chat}~\cite{NExTChat} makes this interface more explicit through \emph{pix2emb}: the LMM generates location embeddings, and task-specific decoders transform those embeddings into boxes or masks~\cite{NExTChat}. \textbf{VistaLLM}~\cite{VistaLLM}, \textbf{Text4Seg}~\cite{Text4Seg}, \textbf{Text4Seg++}~\cite{Text4SegPP}, and \textbf{HiMTok}~\cite{HiMTok} differ mainly in how the structured output is serialized, from point sequences to semantic descriptors and hierarchical mask tokens~\cite{VistaLLM,Text4Seg,Text4SegPP,HiMTok}. Unlike prompt-based families, the MLLM here emits the mask itself or a structured proxy that is later rendered into the final bitmap.

\paragraph{Attention-based prediction.} In this family, the segmentation result is recovered from the model's internal cross-modal attention, which is extracted, converted into mask logits by a lightweight decoder, and optionally refined afterward. \textbf{F-LMM}~\cite{FLMM} is a representative example: it keeps the underlying LMM frozen, extracts stacked word--image attention maps during a normal forward pass, uses a lightweight keyword selector to identify the relevant text spans, feeds the attention stack into a tiny U-Net-like mask head, and then applies a lightweight SAM-based refiner for sharper boundaries. \textbf{DIFFLMM}~\cite{DIFFLMM} follows the same attend-and-segment logic, but strengthens the visual grounding signal by replacing the standard CLIP visual encoder with a diffusion-based one while still relying on weak supervision rather than explicit grounding labels. The main distinction is that the segmentation signal is read out from latent cross-modal attention and only afterward converted into a mask.

\subsubsection{Vision Backbone}

\noindent The choice of vision backbone determines the spatial bandwidth available to the language--pixel interface. If the visual encoder discards too much local structure too early, later language reasoning cannot fully recover precise boundaries. In single-embedding systems such as \textbf{LISA}~\cite{LISA}, \textbf{GLaMM}~\cite{GLaMM}, and \textbf{VideoLISA}~\cite{VideoLISA}, the common pattern is to let the MLLM consume CLIP- or LLaVA-style semantic features while delegating high-resolution detail to SAM or SAM2 features for mask execution. Once grounded dialogue or multimodal reference is added, the encoder must also retain region-level structure rather than only holistic semantics, which is exactly what \textbf{GLaMM}~\cite{GLaMM} and \textbf{AnyRef}~\cite{AnyRef} are designed to exploit. Query-rich models shift more perception burden into the encoder itself: \textbf{GROUNDHOG}~\cite{GROUNDHOG} uses a masked feature extractor to produce visual entity tokens, \textbf{OMG-LLaVA}~\cite{OMGLLaVA} uses a universal segmentation method as visual encoder and supplements it with perception prior embeddings, and \textbf{PSALM}~\cite{PSALM} relies on dense visual features because its mask tokens are decoded internally rather than by an external promptable segmenter. \textbf{PixelLM}~\cite{PixelLM} similarly depends on multi-scale visual features because its lightweight pixel decoder must recover object boundaries without outsourcing the last step to SAM.

The generalist branch makes a different set of backbone choices. \textbf{VistaLLM}~\cite{VistaLLM}, \textbf{VisionLLM v2}~\cite{VisionLLMv2}, and \textbf{HyperSeg}~\cite{HyperSeg} all move more perception burden into the backbone through richer tokenization, fine-grained visual perceivers, or tighter coupling to downstream decoders. In the attention-based branch, \textbf{F-LMM}~\cite{FLMM} keeps the original LMM frozen and mines grounding cues from its native attention maps, whereas \textbf{DIFFLMM}~\cite{DIFFLMM} swaps the standard CLIP encoder for a diffusion-based visual encoder. Once the setting becomes video, temporal memory becomes part of the backbone story itself: \textbf{VideoLISA}~\cite{VideoLISA} balances temporal context and spatial detail through sparse-dense sampling, while \textbf{Sa2VA}~\cite{Sa2VA} inherits SAM2's spatiotemporal memory so that masks can persist consistently across frames.

\subsubsection{Mask Decoder and Output Representation}

The final architectural choice concerns how masks are materialized. The most common executor is still a promptable SAM-family decoder: \textbf{LISA}~\cite{LISA}, \textbf{GLaMM}~\cite{GLaMM}, and \textbf{VideoLISA}~\cite{VideoLISA} all project hidden states into SAM, while \textbf{Sa2VA}~\cite{Sa2VA} upgrades the executor to SAM2 for image/video mask propagation. \textbf{X-SAM}~\cite{XSAM} remains in this SAM-centric lineage, but explicitly revisits the segmentation stack so that one MLLM can cover broader ``any segmentation'' settings. A second route internalizes denser decoders: \textbf{PSALM}~\cite{PSALM} uses mask tokens with a Mask2Former-style generator~\cite{cheng2022masked,li2024box2mask}, \textbf{PixelLM}~\cite{PixelLM} uses a lightweight pixel decoder over a segmentation codebook, \textbf{GROUNDHOG}~\cite{GROUNDHOG} retrieves and merges entity masks from visual entity tokens, and \textbf{OMG-LLaVA}~\cite{OMGLLaVA} keeps one encoder, one decoder, and one LLM for end-to-end pixel reasoning.

The generative line changes the output representation itself rather than only the decoder. \textbf{NExT-Chat}~\cite{NExTChat}, \textbf{Text4Seg++}~\cite{Text4SegPP}, and \textbf{HiMTok}~\cite{HiMTok} illustrate how masks can be serialized as location embeddings, semantic descriptors, or hierarchical tokens. Nearby variants such as \textbf{PaDT}~\cite{PaDT}, \textbf{ARGenSeg}~\cite{ARGenSeg}, and \textbf{UFO}~\cite{UFO} further reduce task-specific machinery.

\subsection{Reliability and Training Strategy}

As models move from description to object-level execution, reliability becomes a first-order concern. \textbf{READ}~\cite{READ}, \textbf{POPEN}~\cite{POPEN}, \textbf{Seg-Zero}~\cite{SegZero}, \textbf{VisionReasoner}~\cite{VisionReasoner}, and \textbf{Seg-R1}~\cite{SegR1} address this issue through token analysis, preference optimization, and RL-style training, while \textbf{PixelThink}~\cite{PixelThink} and \textbf{CoPRS}~\cite{CoPRS} emphasize efficient and spatially interpretable reasoning.

\subsection{Image Referring Expression Segmentation}

In the setting of Image referring expression segmentation, inputs are short expressions referring to a single visible target; recent LMM-based variants allow implicit, knowledge-dependent, or multi-instance expressions. As a result, the boundary between image RES and open-vocabulary dense prediction is becoming less rigid: \textbf{PSALM}\cite{PSALM} and \textbf{UFO}\cite{UFO} extend beyond explicit single-target grounding, with \textbf{LISA}\cite{LISA} and \textbf{GLaMM}\cite{GLaMM} exemplifying this shift. \textbf{GSVA}\cite{GSVA}, \textbf{GROUNDHOG}\cite{GROUNDHOG}, and \textbf{PixelLM}\cite{PixelLM} address this generalized setting via multi-target grounding, no-target rejection, and improved text–mask alignment. Pushing alignment further, \textbf{CoDe}\cite{wu2024image} introduces image–text co-decomposition for fine-grained region–word matching using only text supervision.
Another line of work improves interactivity and reliability. \textbf{SegLLM}~\cite{SegLLM} and \textbf{MIRAS}~\cite{MIRAS} enable multi-round refinement, allowing iterative correction instead of one-shot prediction. \textbf{RSVP}~\cite{RSVP} and \textbf{CoPRS}~\cite{CoPRS} strengthen reasoning–grounding connections, while \textbf{POPEN}~\cite{POPEN}, \textbf{Seg-Zero}~\cite{SegZero}, and \textbf{SegAgent}~\cite{SegAgent} focus on hallucination reduction and alignment. \textbf{PRIMA}~\cite{PRIMA} further extends the instruction-to-mask interface to multi-image reasoning segmentation.

\subsection{Video Referring Expression Segmentation}

Video referring expression segmentation is harder than image RES because the model must jointly solve instruction understanding, temporal localization, and identity preservation across frames. Unlike still images, the correct target may appear only during part of the clip, and once identified must be tracked under motion, occlusion, and viewpoint change. \textbf{VISA}~\cite{VISA} is an early milestone because it introduces the ReasonVOS setting and makes reasoning-heavy video grounding explicit. \textbf{VideoLISA}~\cite{VideoLISA} extends the token-to-mask interface to videos with a \texttt{<TRK>} token for cross-frame segmentation and tracking, while \textbf{Sa2VA}~\cite{Sa2VA} offers a modular design where the MLLM reasons and SAM2 propagates masks.
Later work focuses on stronger temporal representation and unified task design. \textbf{VRS-HQ}~\cite{VRSHQ}, \textbf{GLUS}~\cite{GLUS}, and \textbf{ViLLa}~\cite{ViLLa} improve temporal reasoning and tracking consistency, with \textbf{GLUS}~\cite{GLUS} making explicit the tradeoff between global reasoning on sparse context frames and local tracking on dense query frames. \textbf{HyperSeg}~\cite{HyperSeg}, \textbf{DeSa2VA}~\cite{DeSa2VA}, and \textbf{DecAF}~\cite{DecAF} continue this line through broader task design and improved temporal training.

\subsection{Referring Audio-Visual Segmentation}

Referring audio-visual segmentation extends the referring paradigm beyond language by allowing sound and speech to participate in target specification. The cue may indicate not what an object looks like, but which speech content is associated with it, what sound it emits, or which visual event is synchronized with the audio. \textbf{AnyRef}~\cite{AnyRef} serves as a bridge because it accepts text, boxes, images, and audio as references within one segmentation pipeline. On the benchmark side, \textbf{Ref-AVS}~\cite{RefAVS} establishes a dedicated referring audio-visual segmentation setting with multimodal-cue expressions, while \textbf{OmniAVS}~\cite{OmniAVS} pushes this further with text, speech, sound, and visual cues, including expressions that require complex reasoning and world knowledge.

\subsection{3D Referring Expression Segmentation}

Compared with rapid progress in 2D image, video, and audio-visual settings, 3D referring expression segmentation remains relatively underexplored in current LMM-based research~\cite{MRSegSurvey}. Recent point-cloud-based methods nevertheless show clear evolution in aligning \textbf{multimodal signals, such as language or multi-view imagery,} with 3D geometry. For example, \textbf{MIT}\cite{yang20232d} illustrates how interlaced transformers can implicitly fuse 2D multi-view images and 3D point clouds for segmentation under weak scene-level supervision, laying critical architectural groundwork for multimodal 3D understanding. Focusing specifically on language, \textbf{3D-STMN}\cite{ThreeDSTMN} departs from the traditional proposal-then-matching pipeline by introducing end-to-end superpoint--text matching, arguing that dense superpoint-level alignment is more efficient and faithful than sparse matching.\textbf{RefMask3D}~\cite{RefMask3D} further strengthens this direction with geometry-aware group--word attention and linguistic primitives, emphasizing that 3D referring segmentation is fundamentally a language--geometry fusion task rather than a mask decoding problem.
Meanwhile, the task is becoming more expressive. \textbf{3D-GRES}~\cite{ThreeDGRES} extends the conventional single-target setting to generalized 3D referring expression segmentation, where one expression may correspond to multiple targets. \textbf{3D-DRES}~\cite{ThreeDDRES} further advances the task from sentence-level target identification to phrase-level mapping through the \textbf{DetailRefer} dataset, indicating a shift toward finer-grained compositional grounding. Despite this progress, current 3D methods still rely on point-cloud transformers, superpoint matching, and geometry-aware query design, rather than a mature LMM-native language--mask interface.

\begin{table*}[t]
\centering
\newcommand{\imgicon}{\raisebox{-0.2ex}{\includegraphics[height=3.0ex]{icons/image-icon.png}}}
\newcommand{\vidicon}{\raisebox{-0.2ex}{\includegraphics[height=3.0ex]{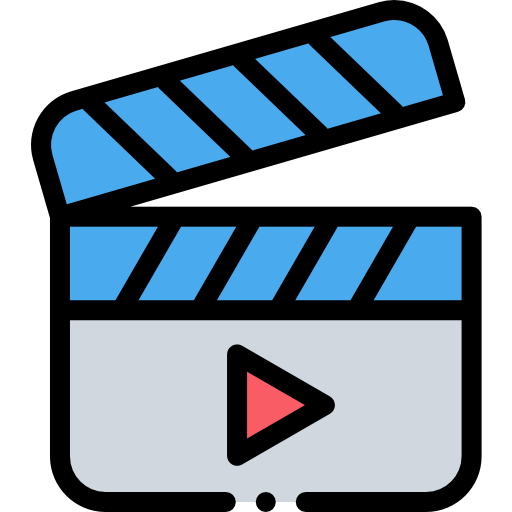}}}
\newcommand{\avicon}{\raisebox{-0.2ex}{\includegraphics[height=3.0ex]{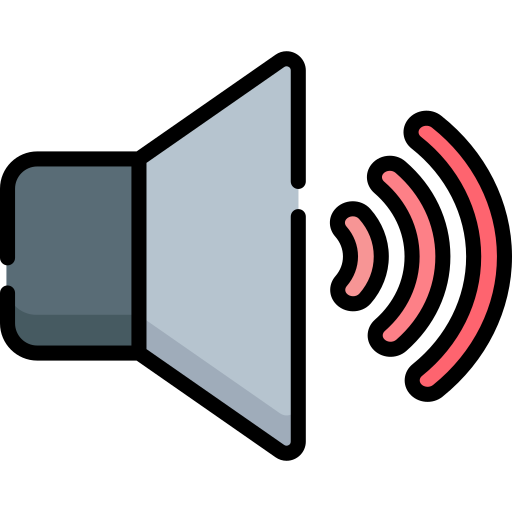}}\hspace{0.15em}\raisebox{-0.2ex}{\includegraphics[height=3.0ex]{icons/video-seg.png}}}
\newcommand{\threedicon}{\raisebox{-0.2ex}{\includegraphics[height=3.0ex]{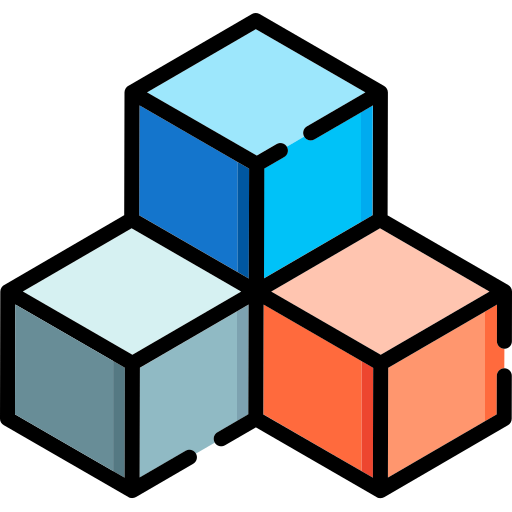}}}
\captionsetup{justification=raggedright, singlelinecheck=false}
\caption{Summary of representative \textbf{Datasets \& Benchmarks} for \textbf{object-centric referring segmentation}.
    \\
    $\bullet$ \textbf{Modality}: Image: \imgicon, Video: \vidicon, Audio-Visual: \avicon, 3D scene: \threedicon.
    \\
    $\bullet$ \textbf{Role}: \bdgDataset: Training Dataset, \bdgBenchmark: Evaluation Benchmark.
}
\vspace{-0.2cm}
\rowcolors{1}{white}{gray!7}
{\renewcommand{\arraystretch}{1.0}
\resizebox{\linewidth}{!}{
\begin{tabular}{llcccl}
    \toprule
    \textbf{Dataset \& Benchmark} & \textbf{Year} & \textbf{Modality} \raisebox{-0.25ex}{\includegraphics[height=2.0ex]{icons/vision.png}} & \textbf{Scale} & \textbf{Role} & \textbf{Task Focus} \raisebox{-0.5ex}{\includegraphics[width=0.021\linewidth]{icons/action.png}}
    \\
    \midrule
    RefCOCO / RefCOCO+ / RefCOCOg~\cite{mao2016generation,nagaraja2016modeling}
    & 2016
    & \imgicon
    & 142K / 142K / 95K
    & \bdgDataset\ \bdgBenchmark
    & Classical referring segmentation and object grounding
    \\

    ReasonSeg~\cite{LISA}
    & 2024
    & \imgicon
    & 1.2K
    & \bdgDataset\ \bdgBenchmark
    & Reasoning segmentation and implicit grounding
    \\
    gRefCOCO~\cite{GRES}
    & 2023
    & \imgicon
    & 278.2K
    & \bdgDataset\ \bdgBenchmark
    & Generalized segmentation with multi- and no-target cases
    \\

    MUSE~\cite{PixelLM}
    & 2023
    & \imgicon
    & 246.0K
    & \bdgDataset\ \bdgBenchmark
    & Open-set multi-target segmentation
    \\
    GranD~\cite{GLaMM}
    & 2024
    & \imgicon
    & 7500K
    & \bdgDataset
    & Grounded concept-mask-text alignment
    \\

    M3G2~\cite{GROUNDHOG}
    & 2024
    & \imgicon
    & 2500K
    & \bdgDataset
    & Multi-grained grounding tuning
    \\
    MRSeg~\cite{SegLLM}
    & 2024
    & \imgicon
    & 1106K
    & \bdgDataset\ \bdgBenchmark
    & Multi-round reasoning segmentation
    \\

    M$^4$Seg~\cite{PRIMA}
    & 2024
    & \imgicon
    & 744.0K
    & \bdgDataset\ \bdgBenchmark
    & Multi-image reasoning segmentation
    \\
    ReasonVOS~\cite{VISA}
    & 2024
    & \vidicon
    & 35.1K
    & \bdgDataset\ \bdgBenchmark
    & Reasoning-based video object segmentation
    \\

    Refer-YouTube-VOS~\cite{ReferYouTubeVOS}
    & 2020
    & \vidicon
    & 27.9K
    & \bdgDataset\ \bdgBenchmark
    & Referring video object segmentation
    \\
    MeViS~\cite{MeViS}
    & 2023
    & \vidicon
    & 28.6K
    & \bdgDataset\ \bdgBenchmark
    & Motion-aware video object segmentation
    \\

    Ref-AVS~\cite{RefAVS}
    & 2024
    & \avicon
    & 20.3K
    & \bdgDataset\ \bdgBenchmark
    & Referring audio-visual segmentation
    \\
    OmniAVS~\cite{OmniAVS}
    & 2025
    & \avicon
    & 59.5K
    & \bdgDataset\ \bdgBenchmark
    & Omnimodal referring audio-visual segmentation
    \\

    DetailRefer~\cite{ThreeDDRES}
    & 2026
    & \threedicon
    & 54.4K
    & \bdgDataset\ \bdgBenchmark
    & Detailed 3D referring segmentation
    \\

    \bottomrule
\end{tabular}}
}
\label{tab:objrefseg-datasets-benchmarks}
\end{table*}

\subsection{Datasets \& Benchmarks}

The evolution of datasets mirrors the evolution of the task. Early LMM-based systems still relied heavily on the \textbf{RefCOCO} family~\cite{mao2016generation,nagaraja2016modeling}, which remains the common substrate for image-level referring segmentation across \textbf{LISA}~\cite{LISA}, \textbf{GLaMM}~\cite{GLaMM}, \textbf{PixelLM}~\cite{PixelLM}, \textbf{GSVA}~\cite{GSVA}, \textbf{AnyRef}~\cite{AnyRef}, \textbf{PSALM}~\cite{PSALM}, and \textbf{GROUNDHOG}~\cite{GROUNDHOG}. Unified open-vocabulary models still lean on dense semantic datasets such as \textbf{ADE20K} and \textbf{Pascal Context}~\cite{ADE20K,PascalContext}. More recent datasets shift emphasis toward reasoning and broader grounding, including \textbf{ReasonSeg}, \textbf{MRSeg}, \textbf{ReasonVOS}, \textbf{Ref-AVS}, and \textbf{DetailRefer}~\cite{LISA,SegLLM,VISA,RefAVS,ThreeDDRES}.

Benchmark design follows the same trajectory. For open-vocabulary and generic dense prediction, \textbf{ADE20K} and \textbf{Pascal Context} remain standard tests~\cite{ADE20K,PascalContext}; in unified LMM systems, these are increasingly evaluated alongside referring segmentation rather than as a separate track~\cite{PSALM,Text4Seg,XSAM,UFO}. For image referring segmentation, the evaluation center has shifted from RefCOCO-style settings~\cite{mao2016generation,nagaraja2016modeling} toward reasoning-heavy and generalized benchmarks such as \textbf{ReasonSeg}, \textbf{MUSE}, and \textbf{MRSeg}~\cite{LISA,PixelLM,SegLLM}. \textbf{M$^4$Seg}, \textbf{ReasonVOS}, and \textbf{Ref-AVS} extend this trajectory to multi-image, video, and audio-visual settings, while \textbf{DetailRefer} marks a more recent 3D extension~\cite{PRIMA,VISA,RefAVS,ThreeDDRES,MRSegSurvey}.
Table~\ref{tab:objrefseg-datasets-benchmarks} summarizes the representative datasets and benchmarks discussed above, organized by modality, role, and task focus.

\section{Object-Centric Visual Editing}
\label{sec5:editing}
The emergence of powerful generative models has revolutionized visual content editing, enabling unprecedented control over image manipulation through intuitive interfaces. This section provides an overview of object-centric editing, spanning diffusion-based approaches to autoregressive models and extending to video and 3D paradigms.

\subsection{Diffusion-based Object Editing}
Latent Diffusion Models (LDMs) have dominated the editing landscape via superior generative priors and ability to model complex image distributions. 


\subsubsection{Attention Control}
Attention manipulation is pivotal for structure-preserving editing~\cite{qi2023fatezero}. \textbf{Prompt-to-Prompt (P2P)} \cite{hertz2022prompt} establishes that cross-attention maps dictate the spatial correspondence between text tokens and image regions. By injecting these maps from the source to target generation, P2P enables mask-free semantic replacement (e.g., ``cat'' $\to$ ``dog'') while preserving the original pose. For non-rigid transformations, \textbf{MasaCtrl} \cite{cao2023masactrl} introduces Mutual Self-Attention to facilitate structural and posture changes while maintaining texture identity. It further utilizes auto-extracted masks from cross-attention to mitigate foreground-background interference.

\subsubsection{Feature Injection}
\textbf{Plug and Play (PnP)} \cite{tumanyan2023plug} enhances attention-based methods by injecting hierarchical spatial features from convolutional layers. By leveraging shallow features for geometry and deep features for semantics, this training-free approach enables text-guided appearance transformations (e.g., sketch-to-image synthesis) while preserving original structural layouts. Advancing beyond feature-level control, \textbf{PixelMan} \cite{jiang2025pixelman} achieves consistent object editing through direct pixel manipulation. By maintaining strict pixel-level correspondence between the source and target images, it guarantees robust structural consistency across multiple editing operations.

\subsubsection{Inversion and Masking}
Real image editing requires accurate reconstruction before modification~\cite{yan2025eedit}. \textbf{Null-Text Inversion} \cite{mokady2023null} addresses DDIM inversion limitations through pivotal inversion and null-text optimization. This technique optimizes the unconditional textual embedding for classifier-free guidance, employing pivotal noise vectors to enhance reconstruction accuracy.
\textbf{LEDITS++} \cite{brack2024ledits++} introduces architecture-agnostic editing through an implicit masking mechanism that extracts semantic regions from cross-attention activations. This eliminates manual mask specification while constraining modifications to relevant areas, supporting multiple simultaneous edits with minimal diffusion steps.
\textbf{FreeFine} \cite{zhu2025training} reconceptualizes geometric editing as a three-stage pipeline: geometric warping, source-hole inpainting, and target-region refinement. This decoupled approach enables precise geometric manipulations without model training.
\textbf{Click2Mask} \cite{regev2025click2mask} simplifies local editing through dynamic mask generation from point-based user input, eliminating manual mask annotation while maintaining precise spatial control.


\subsubsection{Instruction Tuning}
Training-based methods excel when complex reasoning or precise spatial control is required. \textbf{InstructPix2Pix} \cite{brooks2023instructpix2pix} pioneered instruction-following by combining GPT-3 and Stable Diffusion to generate synthetic training data. This framework directly maps input images and textual instructions to edited outputs in a single forward pass.
\textbf{MagicBrush} \cite{zhang2023magicbrush} introduced a substantial manually annotated dataset featuring real images with diverse instruction-output pairs, focusing on object-centric modifications. This human-curated resource improves editing precision compared to synthetic data.
\textbf{Style-Editor} \cite{park2025style} performs object-specific style transfer through specialized loss functions that maintain style consistency within object boundaries while preserving non-target regions.

Further developments include \textbf{AnyEdit} \cite{yu2025anyedit}, which masters unified high-quality image editing for diverse ideas through a single framework, and \textbf{MagicQuill} \cite{liu2025magicquill}, an intelligent interactive system that combines natural language understanding with intuitive interface design for seamless image editing workflows.

\subsubsection{Test-Time Optimization}
\textbf{DragDiffusion} \cite{shi2024dragdiffusion} achieves precise spatial editing through latent space optimization, where diffusion latents are iteratively updated to minimize the distance between control and target points. Supported by fine-tuning for identity preservation, it excels in complex scene manipulation despite high computational overhead. The accompanying \textbf{DragBench} provides a rigorous benchmark for evaluating such point-based interaction paradigms.

\begin{figure*}[t]
\centering
\includegraphics[width=0.96\linewidth]{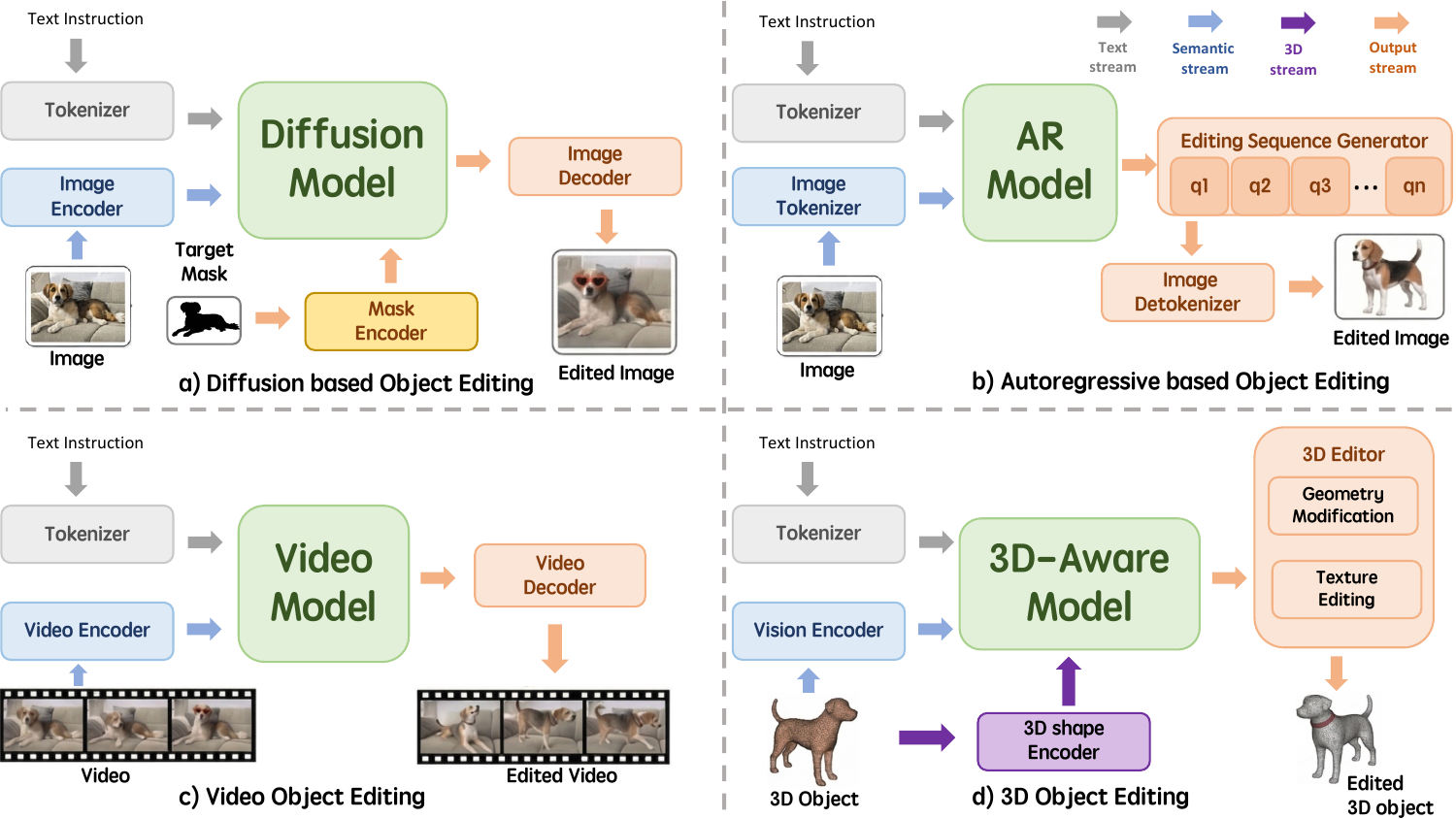}
\caption{Four representative paradigms for generative model-based object editing: (a) Diffusion-based Object Editing, (b) Autoregressive-based Object Editing, (c) Video Object Editing, and (d) 3D Object Editing. They differ in their core generative mechanisms and target modalities, specifically utilizing diffusion processes for standard 2D images, autoregressive models to generate tokenized editing sequences, video models for temporal frame sequences, and 3D-aware models for geometry and texture modifications.}
\label{fig:editing-architecture}
\end{figure*}

\subsection{Autoregressive Object Editing}
Architectural innovations have revitalized autoregressive models for visual editing, addressing limitations in generation speed and editing flexibility while leveraging scaling properties demonstrated in language models.

\subsubsection{Masked and Multi-scale Modeling}
\textbf{MaskGIT} \cite{chang2022maskgit} replaced the traditional raster-scan ordering with a bidirectional transformer that predicts randomly masked tokens. Its iterative refinement process, which generates all tokens simultaneously and then progressively improves them, achieves dramatic acceleration over sequential decoding.
\textbf{Visual Autoregressive (VAR)} \cite{tian2024visual} introduced next-scale prediction, generating images from coarse to fine resolutions rather than token-by-token. This approach demonstrates substantial improvements in both generation quality and inference efficiency. VAR exhibits LLM-like scaling laws and zero-shot generalization to editing tasks, making AR models competitive with diffusion transformers.

\subsubsection{Scale-aware Editing}
\textbf{VAREdit} \cite{mao2025visual} extends VAR's multi-scale framework through scale-aligned reference mechanisms that preserve source image characteristics across different resolution levels. This eliminates complex inversion procedures while preserving fine details.
\textbf{HMAR} \cite{kumbong2025hmar} combines VAR's hierarchical generation with MaskGIT's masking flexibility. Coarse-scale tokens control structure while fine-scale tokens refine textures, supporting both global and localized edits.
\textbf{EditAR} \cite{mu2025editar} unifies segmentation-to-image synthesis, instructed editing, and inpainting under a single conditional generation paradigm, demonstrating the versatility of modern autoregressive architectures.

\subsection{Video Object Editing}
Video editing introduces temporal complexity beyond static image manipulation, requiring modifications to maintain consistency across frames while respecting motion dynamics and temporal coherence~\cite{ye2025unic, ma2025magicstick, bai2025scaling}. The challenge lies in balancing editing flexibility with temporal stability.

\subsubsection{Temporal Propagation}
\textbf{TokenFlow} \cite{geyer2023tokenflow} enforces temporal consistency through diffusion feature manipulation, propagating keyframe features to intermediate frames via linear combination.
\textbf{ContextFlow} \cite{chen2025contextflow} addresses background conflicts through adaptive context enrichment mechanisms that preserve scene coherence during object editing. This facilitates intelligent blending of edited objects with dynamic backgrounds.
\textbf{Anyv2v} \cite{ku2024anyv2v} offers a tuning-free framework for arbitrary video-to-video translations, balancing spatial editing precision with temporal coherence.

\subsubsection{Subject-driven Video Generation}
\textbf{DreamVideo-2} \cite{wei2024dreamvideo} realizes zero-shot subject-driven video customization through attention-based identity preservation and spatial motion guidance, facilitating seamless integration of personalized objects.
\textbf{InsViE-1M} \cite{wu2025insvie} addresses data scarcity by providing an extensive collection of video editing triplets with diverse instructions. This large-scale dataset significantly improves instruction comprehension and motion preservation compared to heuristic-based methods.
\textbf{DIVE} \cite{huang2025dive} tames DINO features for subject-driven video editing, leveraging self-supervised visual representations to maintain subject consistency across frames. \textbf{RACCooN} \cite{yoon2024raccoon} presents a versatile instructional video editing framework with auto-generated narratives, supporting complex multi-step editing operations guided by natural language descriptions.

\subsection{3D Object Editing}
The evolution from implicit neural representations to explicit geometric primitives marks a fundamental shift in 3D editing capabilities~\cite{bai2024real}. This transition reflects the field's growing emphasis on interpretability, efficiency, and direct manipulability of 3D content.

\subsubsection{Implicit NeRF Editing}
\textbf{CLIP-NeRF} \cite{wang2022clip} pioneered text-driven NeRF manipulation through hypernetwork deformation conditioned on CLIP embeddings. However, NeRF's implicit representation entangles objects within the volumetric function, preventing selective manipulation.
\textbf{SPIn-NeRF} \cite{mirzaei2023spin} combines 3D segmentation with 2D inpainting priors across multiple views for object removal and replacement. This method requires retraining the neural field for each edit, limiting interactive applicability.

\begin{center}
    \begin{table*}[t]
\centering
\captionsetup{justification=raggedright, singlelinecheck=false}
\newcommand{\threedicon}{\raisebox{-0.2ex}{\includegraphics[height=3.0ex]{icons/box-seg.png}}}
\caption{Summary of existing \textbf{Datasets \& Benchmarks} for training and evaluating \textbf{object-centric editing methods}.
    \\
    $\bullet$ \textbf{Modality}: \iconImg: Image, \iconVid: Video and \threedicon: 3D scene.
    \\
    $\bullet$ \textbf{Role}: \bdgDataset: Training Dataset, and \bdgBenchmark: Evaluation Benchmark.
}
\vspace{-0.2cm}
\resizebox{\linewidth}{!}{
\begin{tabular}{llcccll}
    \toprule
    \textbf{Dataset \& Benchmark} & \textbf{Year} & \textbf{Modality} \raisebox{-0.25ex}{\includegraphics[width=0.022\linewidth]{icons/vision.png}} & \textbf{Scale} & \textbf{Role} & \textbf{Editing Types} \raisebox{-0.5ex}{\includegraphics[width=0.021\linewidth]{icons/action.png}} & \textbf{Evaluation Metrics}
    \\
    \midrule
    \midrule
    MagicBrush \cite{zhang2023magicbrush}
    & 2023
    & \iconImg
    & 10K triplets
    & \bdgDataset\ \bdgBenchmark
    & \makecell[l]{Add, Remove, Replace, \\ Color, Action, Pattern}
    & L1, L2, CLIP-I/T, DINO
    \\
    \rowcolor{gray!7}
    TEdBench++ \cite{brack2024ledits++}
    & 2024
    & \iconImg
    & 120 entries
    & \bdgBenchmark
    & \makecell[l]{Multi-conditioning, Removal, \\ Style Transfer, Complex Replace}
    & Success Rate, LPIPS
    \\
    DragBench \cite{shi2024dragdiffusion}
    & 2024
    & \iconImg
    & 205 images
    & \bdgBenchmark
    & Drag / Point-based
    & Mean Distance, Image Fidelity
    \\
    \rowcolor{gray!7}
    InsViE-1M \cite{wu2025insvie}
    & 2025
    & \iconVid
    & 1.02M triplets
    & \bdgDataset\ \bdgBenchmark
    & \makecell[l]{Substitution, Color, Style, \\ Add, Remove, Weather}
    & \makecell[l]{CLIP-T, PickScore, \\ DOVER, Flow EPE}
    \\
    EdiVal-Agent \cite{chen2025edival}
    & 2025
    & \iconImg
    & \makecell[c]{572 images, \\ 1,716 instructions}
    & \bdgBenchmark
    & \makecell[l]{Subject Add/Remove/Replace, \\ Color, Material, Position, Count}
    & \makecell[l]{EdiVal-IF, EdiVal-CC, \\ EdiVal-VQ, EdiVal-O}
    \\
    \rowcolor{gray!7}
    OBJECT \cite{michel2023object}
    & 2023
    & \threedicon
    & 400K triplets
    & \bdgDataset\ \bdgBenchmark
    & \makecell[l]{Translation, Rotation, \ Insertion, Removal}
    & PSNR, SSIM, LPIPS, FID
    \\
    SPIn-NeRF \cite{mirzaei2023spin}
    & 2023
    & \threedicon
    & 10 scenes
    & \bdgBenchmark
    & Object removal, Inpainting
    & LPIPS, FID
    \\
    \rowcolor{gray!7}
    PhysGaia \cite{kim2025physgaia}
    & 2025
    & \threedicon
    & 17 dynamic scenes
    & \bdgBenchmark
    & \makecell[l]{Physical simulation, \ Multi-body dynamics}
    & FID, KID, Dynamic Score
    \\
    \bottomrule
\end{tabular}}
\label{tab:editing_benchmark}
\end{table*}
\end{center}

\subsubsection{Explicit 3D Gaussian Splatting (3DGS)}
3D Gaussian Splatting (3DGS) enables object-centric editing via explicit scene representations composed of discrete Gaussian primitives with rich attributes. 
\textbf{GaussianGrouping} \cite{ye2024gaussian} lifts 2D segmentation to 3D by assigning semantic labels to Gaussians, enabling intuitive object selection and rigid transformations. 
\textbf{PhysGaussian} \cite{xie2024physgaussian} introduces physical priors, modeling Gaussians as particles to support realistic deformation and dynamics. 
\textbf{Drag Your Gaussian} \cite{qu2025drag} extends drag-based interaction to 3D through iterative optimization of Gaussian parameters with score distillation constraints. 
\textbf{GaussianEditor} \cite{chen2024gaussianeditor} supports semantic and geometric editing, combining efficient manipulation with generative guidance for texture transfer and scene composition. 
Subsequent work includes \textbf{Localized Gaussian Splatting Editing} \cite{xiao2025localized} for context-aware local edits, \textbf{Mani-GS} \cite{gao2025mani} for more precise manipulation via mesh priors, and \textbf{DriveEditor} \cite{liang2025driveeditor}, which provides a unified 3D-guided framework for controllable object editing in driving scenes.

\subsection{Datasets and Benchmarks}
Standardized benchmarks enable systematic evaluation of editing methods. \textbf{MagicBrush} \cite{zhang2023magicbrush} offers manually annotated images with diverse editing instructions, while \textbf{TEdBench++} \cite{brack2024ledits++} provides comprehensive evaluation for text-based editing. \textbf{DragBench} \cite{shi2024dragdiffusion} focuses on interactive point-based editing, and \textbf{InsViE-1M} \cite{wu2025insvie} supports large-scale video editing evaluation. 
\textbf{EdiVal-Agent} \cite{chen2025edival} introduces an object-centric framework for fine-grained, multi-turn editing assessment.
Evaluation has also expanded to 3D and physical domains. \textbf{OBJECT} \cite{michel2023object} is a large-scale dataset for 3D-aware object editing with spatial operations, while \textbf{SPIn-NeRF} \cite{mirzaei2023spin} targets object removal and inpainting in real scenes. \textbf{PhysGaia} \cite{kim2025physgaia} further benchmarks physics-aware interactions in dynamic environments. Together, these benchmarks support fair comparison and highlight remaining challenges.

\section{Object-Centric Visual Generation}
\label{sec6:generation}

The generative landscape across images, videos, and 3D is shifting from holistic synthesis to object-centric modeling. Modern LMMs span diffusion and autoregressive image models, temporal video generation, and 3D representations such as 3DGS, enabling unified representation and manipulation of entities across space and time.

\begin{figure*}[t]
\centering
\includegraphics[width=0.98\linewidth]{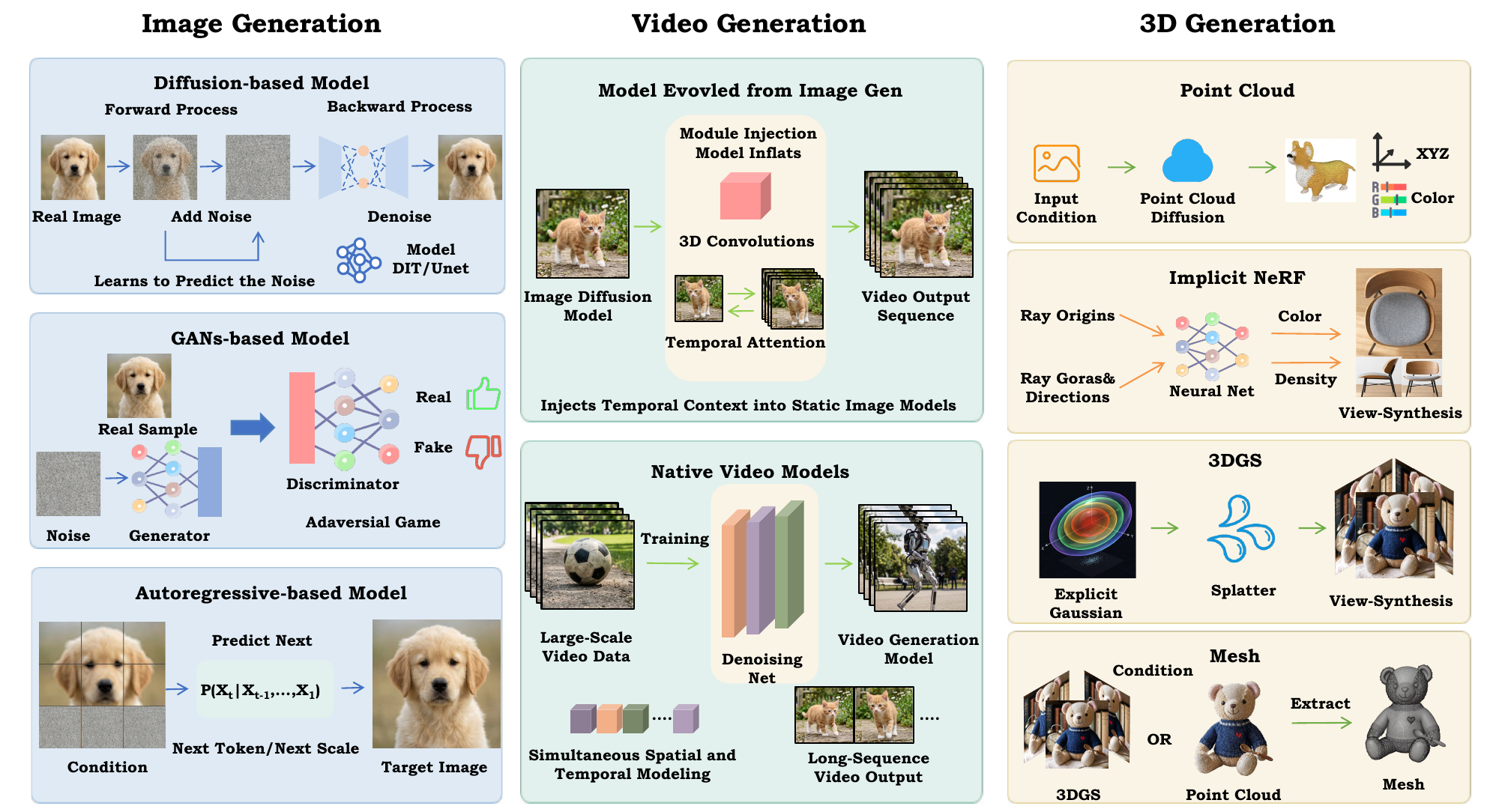}
\caption{\textbf{Overview of generative methodologies for multimodal content creation.} The figure categorizes mainstream approaches for image, video, and 3D generation, highlighting the evolution from static 2D synthesis to complex spatial and temporal modeling.}
\label{fig:generation-architecture}
\end{figure*}
\subsection{Image Generation}
\subsubsection{Autoregressive-based Image Generation}
Autoregressive (AR) models treat image generation as a sequential prediction task, offering tractable likelihood estimation but traditionally suffering from high computational costs during inference. Key advancements in this paradigm include \textbf{VQ-VAE}~\cite{van2017neural}, which mitigates these costs by representing image patches as sequences of discrete tokens using a learned codebook; \textbf{LlamaGen}~\cite{sun2024autoregressive}, which demonstrates that properly scaled vanilla Transformers can achieve highly competitive performance using the standard ``next-token prediction'' approach; and \textbf{Visual Autoregressive Modeling (VAR)}~\cite{tian2024visual}, which introduces a coarse-to-fine ``next-scale prediction'' paradigm that resolves bidirectional dependency issues, ultimately allowing AR models to surpass diffusion models in both image quality and inference speed.

\subsubsection{GANs-based Image Generation}
Generative Adversarial Networks (GANs) utilize a deepfake detection game between a generator and a discriminator to produce high-fidelity images. \textbf{Image Generation from Layout}~\cite{zhao2019image} uses graph convolutional networks to transform user-defined bounding boxes into structured scenes, providing precise spatial control over object placement. \textbf{Obj-GAN}~\cite{li2019object} first predicts a layout from a text prompt and then employs object-aware discriminators to ensure each individual element is semantically accurate. Together, these methods shifted AI synthesis from ``blurry whole-image generation'' to organized, object-level scene construction. More recent work \textbf{Aurora}~\cite{zhu2025exploring} integrates MoE mechanism to enlarge the generator's capacity without a heavy computational burden.

\subsubsection{Diffusion-based Image Generation}
Diffusion models surpass GANs in stability and diversity by reversing a noise process. Key advances include \textbf{DDPM}~\cite{ho2020denoising} for scalable training, \textbf{DDIM}~\cite{song2020denoising} for faster sampling, and \textbf{LDM}~\cite{rombach2022high} for efficient latent-space diffusion. 
Building on these, methods such as \textbf{Instance Diffusion}~\cite{wang2024instancediffusion} and \textbf{LayoutDiffusion}~\cite{zheng2023layoutdiffusion} enable layout-controlled generation, while \textbf{Follow-Your-Color}~\cite{zhang2025follow} supports reference-guided colorization. For customization, \textbf{Textual Inversion}~\cite{gal2022image}, \textbf{DreamBooth}~\cite{ruiz2023dreambooth}, and \textbf{Custom Diffusion}~\cite{kumari2023multi} progressively improve efficiency and multi-concept learning, with \textbf{MUDI}~\cite{jang2024identity} further enhancing initialization via SAM and box guidance. 
Recently, Transformer-based diffusion (DiT) marks a shift from U-Net architectures. \textbf{FLUX}~\cite{blackforestlabs2024flux1dev} adopts a scalable Transformer with flow matching for faster convergence and better prompt alignment. Building on this, \textbf{Flux Already Knows}~\cite{kang2025flux} eliminates training by arranging reference images in latent space, leveraging the model’s inherent ability to generalize object identity.

\subsection{Video Generation}
Video generation has experienced explosive growth, driven by advancements in diffusion models and transformers. The primary challenge lies in maintaining spatial fidelity across frames while generating temporally consistent videos.
\subsubsection{Models Evolved from Image Generation}
This approach leverages the massive visual knowledge already encoded in pre-trained Text-to-Image models (like Stable Diffusion). By injecting temporal dynamics into a frozen or lightly fine-tuned image model, researchers can generate video without needing to train a visual foundation model from scratch.
\textbf{AnimateDiff}~\cite{guo2023animatediff} trains a motion module which can be plugged into personalized Stable Diffusion to instantly turn it into an animation model. \textbf{Tune A Video}~\cite{wu2023tune} inflates a T2I model and fine-tunes it on just a single input video, allowing it to learn the specific motion and apply it to new text prompts. \textbf{Stable Video Diffusion}~\cite{blattmann2023stable} which evolves from Stable Diffusion 2.1, is trained by first pre-training on images, then establishes motion capabilities by temporal layers trained on videos. \textbf{VideoCrafter}~\cite{chen2023videocrafter1} builds upon SD 2.1 by incorporating temporal attention layers into the SD UNet and prompts I2V generation with text-aligned rich image embedding.
\subsubsection{Native Video Models Trained from Scratch}
Instead of relying on image priors, these models are trained from the ground up on massive, curated datasets of video-text pairs. This allows the model to deeply learn real-world physics, object permanence, and complex spatial-temporal relationships. Then the model can generate longer videos with superior temporal consistency and develop a rudimentary world model. \textbf{Sora}~\cite{brooks2024video} trains on, understands, and generates videos and images at
their native sizes. It uses spacetime patches to generate up to 60 seconds of highly coherent, photorealistic video with complex camera motions and consistent physics. \textbf{Veo}~\cite{wiedemer2025video} prompts the model to form its deep semantic understanding of prompts, capable of handling complex cinematic requests. \textbf{Wan}~\cite{wan2025wan} is a comprehensive, open-source suite of video foundation models which leverage 3D causal VAE and scalable pre-training strategies achieving comparable performance with those closed commercial solutions. Based on Wan-2.1, \textbf{Follow-Your-Motion}~\cite{ma2025follow} proposes a three-stage fine-tuning strategy that decouples the spatial appearance and temporal dynamics heads to apply the motion to novel objects.
\subsection{3D Object Generation}
The development of modern 3D generative AI is best categorized by its underlying 3D representation methods. The choice of representation governs a model's architecture, optimization techniques, rendering efficiency, and its ultimate suitability for downstream applications. There are four primary representational paradigms:
\subsubsection{Point Cloud}
Point clouds are the most fundamental 3D representation. They contain coordinates (x, y, z) alongside color or normal features. \textbf{Point-E}~\cite{nichol2022point} breaks the generation process into three steps: text to image generation, image to coarse point cloud, upsampling to high-resolution point cloud. \textbf{LION}~\cite{vahdat2022lion} compresses point clouds into a latent space for diffusion generation, improving both generation quality and shape coherence. \textbf{GPT4Point}~\cite{qi2024gpt4point} utilizes Q-Former to attain text-aligned point cloud embeddings, and then improves the generation quality.
\subsubsection{Implicit NeRF}
Implicit representations do not explicitly store points or faces. Instead, they use neural networks to map continuous 3D spatial coordinates (x, y, z) and viewing directions to the volume density and color at that location. \textbf{DreamFusion}~\cite{poole2022dreamfusion}: pioneered the use of 2D diffusion model priors (SDS Loss) to optimize a NeRF, achieving zero-shot text-to-3D generation.
\textbf{ProlificDreamer}~\cite{wang2023prolificdreamer} proposed the VSD (Variational Score Distillation) loss function, solving the over-saturation and over-smoothing issues found in DreamFusion to generate incredibly high-resolution NeRFs.
\subsubsection{3DGS}
It represents scenes using thousands of ``3D Gaussian blobs'', each possessing its own covariance (shape/size), opacity, and color, and heavily improves the speed of rendering.
\textbf{DreamGaussian}~\cite{tang2023dreamgaussian} introduced 3DGS into the text/image-to-3D generation framework, slashing optimization time drastically. 
\textbf{LGM}~\cite{tang2024lgm} efficiently predicts multi-view 3D Gaussians from a single image using multi-view diffusion models and U-Net architecture. To overcome the distortion and blurriness caused by geometric ambiguity and unstructured Gaussians, \textbf{GS-RGBN}~\cite{shen2025high} introduces a hybrid Voxel-Gaussian model, utilizing a feature-level cross-volume fusion module to effectively align semantic and geometric cues from multi-view RGB and normal images.
\textbf{SAM3D}~\cite{chen2025sam} builds a MITL pipeline to curate data at unprecedented scale. Then it introduces a new foundation model for single-view 3D generation via LLM-style pretraining and post-training on the data.

\begin{center}
    \begin{table*}[t]
\centering
\captionsetup{justification=raggedright, singlelinecheck=false}
\caption{Summary of existing \textbf{Datasets \& Benchmarks} for training and evaluating \textbf{image/video/3D generation methods}.
    \\
    $\bullet$ \textbf{Role}: \bdgDataset: Training Dataset, and \bdgBenchmark: Evaluation Benchmark.
}
\vspace{-0.2cm}
\rowcolors{1}{white}{gray!7}
\resizebox{\linewidth}{!}{
\begin{tabular}{llccll}
    \toprule
    \textbf{Benchmark} & \textbf{Year} & \textbf{Scale} & \textbf{Role} & \textbf{Data Components} \raisebox{-0.5ex}{\includegraphics[width=0.021\linewidth]{icons/action.png}} & \textbf{Usage}
    \\
    \midrule
    \midrule
    LAION-400M \cite{schuhmann2021laion}
    & 2021
    & 400M pairs
    & \bdgDataset
    & Image-Text pairs
    & Text-guided image generation training
    \\
    LAION-5B \cite{schuhmann2022laion}
    & 2022
    & 5B pairs
    & \bdgDataset
    & Image-Text pairs
    & Text-guided image generation training
    \\
    MS-COCO \cite{lin2014microsoft}
    & 2014
    & 328k images
    & \bdgBenchmark
    & Images, Bbox, Captions, Mask
    & Generation evaluation with FID, CLIP
    \\
    InternVid \cite{wang2023internvid}
    & 2023
    & \makecell[c]{7M videos, 243M clips}
    & \bdgDataset
    & Video-Text pairs
    & Text-driven video generation
    \\
    Panda-70M \cite{chen2024panda}
    & 2024
    & 70M videos
    & \bdgDataset
    & Video-Text pairs
    & Text-driven video generation
    \\
    Text2Shape \cite{chen2018text2shape}
    & 2018
    & \makecell[c]{75K paris}
    & \bdgDataset
    & Captions, Colored Voxels
    & \makecell[l]{Conditional 3D generation}
    \\
    Gobjaverse \cite{zuo2024sparse3d}
    & 2024
    & \makecell[c]{300M rendered images}
    & \bdgDataset
    & Albedo, RGB, Depth, Normal map
    & Training 3DGS, NeRFs
    \\
    Objaverse-XL \cite{deitke2023objaverse-xl}
    & 2023
    & \makecell[c]{10M 3D objects}
    & \bdgDataset
    & 3D assets
    & \makecell[l]{Novel-view Synthesis, \\ 3D object generation}
    \\
    Objaverse \cite{deitke2023objaverse}
    & 2022
    & 800K 3D models
    & \bdgDataset
    & 3D assets with descriptive captions
    & Training generative 3D models
    \\
    OmniObject3D \cite{wu2023omniobject3d}
    & 2023
    & \makecell[c]{6,000 objects}
    & \bdgDataset\ \bdgBenchmark
    & \makecell[l]{Textured Meshes, Point Clouds, \\ Multi-view Rendered Images, \\ Multiple Real-captured Videos}
    & \makecell[l]{Novel-view synthesis, \\ 3D object generation, \\ Neural surface reconstruction}
    \\
    \bottomrule
\end{tabular}}
\label{tab:editing_benchmark}
\end{table*}
\end{center}

\subsubsection{Mesh}
Generating meshes directly from scratch is incredibly difficult because their topological structure is irregular. Some methods such as \textbf{Magic3D}~\cite{lin2023magic3d}, \textbf{DreamGaussian}~\cite{tang2023dreamgaussian}, usually adopt a hybrid pipeline: they first generate coarse representation (like point clouds or 3DGS), and then extract it as a mesh, while \textbf{Unilat3D}~\cite{wu2025unilat3d} contends that this decoupled approach is prone to geometry-texture inconsistencies and thus formulates an end-to-end framework where geometry and texture features are unified within a shared latent space, facilitating an one-stage generation.
Recently, some unified models like \textbf{ShapeLLM-Omni}~\cite{ye2025shapellm} have evolved to enhance 3D mesh generation by incorporating MLLM which has a strong ability to understand input requests.

\subsection{Datasets and Benchmarks}

For image generation, \textbf{LAION-400M}~\cite{schuhmann2021laion} and its larger successor \textbf{LAION-5B}~\cite{schuhmann2022laion} serve as key training corpora for models such as Stable Diffusion. Despite its smaller scale, \textbf{MS-COCO}~\cite{lin2014microsoft} remains a standard benchmark for text-to-image alignment due to its high-quality human annotations.
For video generation, \textbf{Panda-70M}~\cite{chen2024panda} provides 70 million high-definition videos with strong semantic alignment, while \textbf{InternVid}~\cite{wang2023internvid} uses LLM-based curation to construct a larger collection of 234 million video--text pairs.
For 3D generation, \textbf{Text2Shape}~\cite{chen2018text2shape} is an early benchmark with human-annotated descriptions for object categories like chairs and tables. \textbf{Objaverse-XL}~\cite{deitke2023objaverse-xl} scales this space to over 10 million deduplicated 3D objects. Built on \textbf{Objaverse}~\cite{deitke2023objaverse}, \textbf{Gobjaverse}~\cite{zuo2024sparse3d} provides high-resolution, background-free multi-view renders for training generative 3DGS and NeRF models. \textbf{OmniObject3D}~\cite{wu2023omniobject3d} further contributes high-quality real-world scans with meshes, point clouds, multi-view images, and videos across 190 categories.
\section{Future Direction}
\label{sec7:future}
The evolutionary trajectory of Multimodal Large Language Models (MLLMs) is shifting from passive 2D image observers to active agents in continuous 3D and temporal environments. As the field approaches the limits of simple parameter scaling, advancing robust, pixel-centric multimodal systems requires overcoming structural and cognitive bottlenecks. We delineate five strategic research frontiers:

\textbf{Unified Representation and Task-Agnostic Architectures. }
Current fragmented pipelines (e.g., separate segmentation heads) must evolve into fully unified, end-to-end frameworks. Inspired by recent works like Sa2VA\cite{Sa2VA}, the paradigm is shifting toward ``understanding as segmentation''. Future MLLMs should map continuous visual features and discrete text into a shared, highly aligned latent space, executing complex visual tasks via pure sequence-to-sequence token generation. This fundamentally eliminates task-specific heads and mitigates cross-modal interference.

\textbf{Long-term Spatio-temporal Consistency in Dynamic Scenes}
Maintaining \textit{instance permanence} across unstructured video sequences remains a cognitive bottleneck. Transcending simple sliding-window mechanisms requires integrating \textit{Spatial Anchor-based Chain-of-Thought (CoT)} reasoning to track objects and logically deduce dynamic scenes through complex occlusions. Furthermore, resolving temporal flickering in generative editing demands the native injection of physical motion priors and dynamic memory banks (e.g., token-compressing queues) into temporal attention layers.

\textbf{Fine-grained Controllability and Attribute Disentanglement.} 
To bridge the gap between high-level LLM reasoning and low-level pixel manipulation, models need highly compositional visual tokens that explicitly decouple shape, appearance, illumination, and kinematics. Establishing a robust ``Chain-of-Spatial-Reasoning'' will enable models to map multi-turn instructions to precise, localized modifications, preventing semantic bleeding and preserving non-target regions during complex visual synthesis.

\textbf{Scaling with Weakly-supervised and Synthetic Data Engines.} 
Given the prohibitive cost of dense mask-level annotations, training paradigms must shift toward automated, self-sustaining data engines. Leveraging foundational models as self-labelers, employing \textit{LLM-as-a-Judge} frameworks for data curation, and harnessing high-fidelity synthetic data (3D simulations and diffusion models) are highly scalable strategies. However, robust optimization algorithms are required to bridge the ``sim-to-real'' domain gap and mitigate auto-generated hallucination noise.

\textbf{Towards Embodied AI and Real-time Active Perception.} 
The ultimate frontier for pixel-centric MLLMs is Embodied AI, demanding continuous interaction in real-world environments. This paradigm shift introduces four key challenges:

\begin{itemize}
\item \textbf{Egocentric Adaptation:} Maintaining robust spatial awareness and object tracking under chaotic, first-person visual streams characterized by motion blur and rapid viewpoint shifts.
\item \textbf{Affordance Grounding:} Moving beyond 2D bounding boxes to pixel-precise affordance understanding, enabling agents to deduce not only ``what'' an object is, but ``where'' and ``how'' to physically interact with it.
\item \textbf{Reinforcement Learning (RL) Integration:} Aligning visual-language policies with physical laws and complex sequential goals via reinforcement learning algorithms, 
while utilizing \textit{LLM-as-a-Judge} mechanisms to dynamically synthesize fine-grained shaped rewards.
\item \textbf{Real-time Edge Deployment:} Achieving low-latency inference on resource-constrained platforms through algorithm–hardware co-design is essential for enabling MLLMs as cognitive engines in embodied systems.
\end{itemize}

\section{Conclusion}
\label{sec8:conclusion}
This paper examined the emerging intersection between large multimodal models and object-centric vision, highlighting a unified perspective spanning object understanding, segmentation, editing, and generation. We showed that, although recent advances have substantially enhanced multimodal reasoning and controllability, existing systems still face persistent challenges in precise grounding, spatial reasoning, and consistent object-level interaction. A notable trend is the movement toward unified object-centric frameworks that bridge perception and generation through shared representations. Looking ahead, improving grounding accuracy, enabling fine-grained and cross-modal controllability, and establishing robust evaluation protocols will be crucial for building reliable, scalable, and generalizable object-centric multimodal systems.

\bibliographystyle{plainnat}
\bibliography{main}

\end{document}